\documentclass{applemlr}

\usepackage{amsmath}
\usepackage{enumerate}
\usepackage{algorithm}
\usepackage{algpseudocode}
\usepackage{amsfonts}
\usepackage{amsthm}
\usepackage{cleveref}
\usepackage{diagbox}
\usepackage{colortbl}
\usepackage{amssymb}
\usepackage{xspace}
\usepackage{wrapfig}
\usepackage{adjustbox}
\usepackage{tabularx}
\usepackage{booktabs}
\usepackage{mathtools}
\usepackage{tikz}
\usepackage{enumitem}
\usepackage{silence}
\usepackage{dsfont}
\usepackage[table]{xcolor}
\usepackage[dvipsnames]{xcolor}
\usepackage{multirow}
\usepackage{makecell}
\usepackage{xfakebold}
\usepackage[most]{tcolorbox}
\tcbuselibrary{listings, breakable}

\usepackage{amsmath,amsfonts,bm}

\def\eqref#1{equation~\ref{#1}}

\def\1{\bm{1}}

\DeclareMathAlphabet{\mathsfit}{\encodingdefault}{\sfdefault}{m}{sl}
\SetMathAlphabet{\mathsfit}{bold}{\encodingdefault}{\sfdefault}{bx}{n}

\definecolor{textgray}{HTML}{6E6E73}
\usetikzlibrary{positioning, calc}
\usetikzlibrary{decorations.pathmorphing}

\makeatletter
\patchcmd{\wrong@fontshape}{\@gobbletwo}{}{}{}
\makeatother
\numberwithin{equation}{section}
\makeatletter
\AtBeginDocument{
  \urlstyle{sf}
  
}
\makeatother

\definecolor{light}{RGB}{125, 125, 125}
\crefname{tcb@cnt@pbox}{code}{code}
\Crefname{tcb@cnt@pbox}{Code}{Code}
\crefname{assumption}{assumption}{assumption}
\Crefname{assumption}{Assumption}{Assumptions}

\newtcolorbox[auto counter]{pbox}[2][]{
  colback=white,
  title=Code~\thetcbcounter: #2,
  #1,fonttitle=\sffamily,
  fontupper=\sffamily,
  arc=2pt,
  colframe=bgcolor,
  coltitle=fgcolor,
  colbacktitle=bgcolor,
  toptitle=0.25cm,
  bottomtitle=0.125cm
}

\makeatletter
\newcommand\applefootnote[1]{%
  \begingroup
  \renewcommand\thefootnote{}%
  \renewcommand\@makefntext[1]{\noindent##1}%
  \footnote{#1}%
  \addtocounter{footnote}{-1}%
  \endgroup
}
\makeatother

\definecolor{cverbbg}{gray}{0.90}

\crefname{section}{Sec.}{Secs.}
\crefname{figure}{Fig.}{Figs.}
\crefname{table}{Table}{Tables}
\crefname{lstlisting}{listing}{listings}
\crefname{equation}{Eq.}{Eqs.}

\crefname{appendix}{Appendix}{Appendices}
\Crefname{appendix}{Appendix}{Appendices}
\crefname{subappendix}{Appendix}{Appendices}
\Crefname{subappendix}{Appendix}{Appendices}
\crefname{subsubappendix}{Appendix}{Appendices}
\Crefname{subsubappendix}{Appendix}{Appendices}

\newtcbinputlisting{\promptbox}[3][gray]{
  listing file={#3}, listing only, breakable,
  colback=#1!5, colframe=#1!60,
  fonttitle=\bfseries, title={#2},
  listing options={basicstyle=\small\ttfamily,
                   breaklines=true, columns=fullflexible}
}

\newcommand{\promptfigure}[5][gray]{%
  \promptbox[#1]{#2}{#3}%
  \nopagebreak                 
  \captionof{figure}{#4}\label{#5}%
  \bigskip
}

\newcommand{\ours}{JudgeMix\xspace}

\title{JudgeMix: Judge-Guided Mixed-Rollout Data Synthesis for LLM Agents}

\author{Cheng-Yu Hsieh}
\author{Ting-Yao Hu}
\author{Sanjoy Chowdhury}
\author{Fatih Ilhan}
\author{Oncel Tuzel}
\author{Raviteja Vemulapalli}

\affiliation{Apple}

\abstract{
Supervised fine-tuning (SFT) on expert trajectories is a standard paradigm for training LLM agents. However, it suffers from a well-known training-inference distribution mismatch, calling for target-aware data synthesis strategies that ground the training data in the target model's native state distribution.
To bridge this gap, emerging approaches incorporate the expert either to guide mixed rollouts alongside the target model, or to provide dense step-wise supervision.
Nonetheless, existing methods apply expert assistance blind to the target model's inherent behavior: they either disrupt the rollout using sub-optimal random intervention heuristics, or they enforce indiscriminate uniform supervision that unnecessarily overwrites the model's already-correct actions.
We introduce JudgeMix, a judge-guided data synthesis framework that dynamically interleaves the target model and expert based on step-wise semantic evaluations. JudgeMix queries expert guidance precisely when the target model struggles, ensuring high trajectory success rates while keeping the state distribution aligned with the target model. It additionally uses the judge evaluation to reweight expert supervision on different steps, preserving model's inherent valid behavior while correcting its errors.
On two agentic tool-use benchmarks, AppWorld and AutomationBench, JudgeMix demonstrates consistent superior performance to existing methods, with improvements up to 10 pass@1 points.
}

\metadata[Correspondence]{\sffamily Cheng-Yu Hsieh: \url{cydhsieh@apple.com}}

\begin{document}

\maketitle

\section{Introduction}
Autonomous LLM agents have enabled a wide variety of new AI applications by interacting with complex environments to complete multi-step tasks~\citep{yao2022react,zhou2024webarena,xie2024osworld,merrill2026terminalbench}.
In our daily lives, these agents serve as personal assistants for tasks like email and calendar management~\citep{trivedi2024appworld,yao2024tau,xiu2026astra}; in the workplace, they automate enterprise workflows and operate specialized software~\citep{yang2024swe,drouin2024workarena,shepard2026automationbench}.
To develop these capabilities, a standard practice is to train LLMs on agentic trajectory datasets with supervised fine-tuning (SFT)~\citep{zeng2024agenttuning,qin2024toolllm,chen-etal-2024-agent,raoof2026openthoughts,blakeman2026nemotron}.
These trajectory datasets form the foundation of agentic capabilities, teaching the model how to reason across intermediate steps, utilize appropriate tools, responding to environment feedback, and navigate complex environments to complete a task.


The standard practice for agentic SFT dataset curation is to collect golden trajectories exclusively from an expert~\citep{qin2024toolllm,song-etal-2024-agentbank,yang2026swe}. While this yields high-quality demonstration data, the curation process is agnostic to the capabilities of the specific target model to be fine-tuned.
Thus, this model-agnostic data generation approach leads to a well-known distribution mismatch between the expert's training states and the target model's inference-time rollouts, leaving the target model brittle once it inevitably deviates from the ideal trajectory~\citep{ross2011reduction,bengio2015scheduled,agarwal2024policy}.

This limitation motivates an emerging direction of target-aware agentic data curation, which aims to build datasets grounded in the target model's actual state distribution~\citep{yuan2023scaling,lauffer2025imitation,sun2026step}.
To achieve this goal while still benefiting from the expert knowledge, recent works generally incorporate both the target model and the expert into the data generation process, utilizing the expert in one of two ways (or a combination thereof): to actively guide the trajectory rollout~\citep{lauffer2025imitation,li2026policy,da2025agent}, or to provide dense target supervision~\citep{agarwal2024policy,sun2026step}.
In the former, the expert generates some steps within the rollout to keep the trajectory from failing. In the latter, the expert acts as an oracle to provide target training labels at every step.

However, these approaches apply expert assistance blindly, ignoring the target model's inherent behavior during the rollout. Specifically, when the expert is used to guide the trajectory, existing works rely on arbitrary heuristics to determine on which steps the expert intervenes. Without consideration of the actual rollout semantics, random intervention could introduce unnecessary disruption of target model's native generation, or be too late to recover from failures.
On the other hand, when the expert is used to provide supervision, current works indiscriminately apply a uniform learning weight across all rollout states. This forces the target model to mimic the expert's actions even at states where its own native generations are valid, imposing unnecessary behavioral changes.

In this work, we introduce \ours, a judge-guided data generation framework for constructing target-aware agentic SFT datasets. The core idea is to employ a judge that evaluates the semantic correctness of the target model's proposed action at every step within a rollout. This step-wise evaluation provides valuable signal on the target model's competence at different states, which \ours exploits to drive three core improvements.
First, \ours utilizes this signal to strategically determine expert interventions: expert guidance happens \textit{only} when the target model struggles.
This mechanism ensures precise expert assistance at necessary steps, enabling high trajectory success rate while keeping the state distribution aligned with the target model.
Second, by pinpointing the exact states where the target model fails, \ours naturally collects supervision that explicitly teaches the model how to recover from its own errors.
Finally, by distinguishing between the target model's successes and failures, \ours dynamically reweights the expert supervision during fine-tuning. This encourages the retention of the model's valid, native behaviors while providing supervision to rectify its errors.

On two complex agentic tool-use benchmarks, AppWorld~\citep{trivedi2024appworld} and AutomationBench~\citep{shepard2026automationbench}, \ours demonstrates consistent improvements over existing approaches across different target models with varying competence.
Compared to the standard model-agnostic data generation approach, \ours shows up to 8 and 10 points improvements (pass@1) on AppWorld and AutomationBench, respectively.

\section{Preliminaries}
\label{sec:preliminaries}

\subsection{Problem setup}
We consider an agentic task as a sequence of interactions between an agent and an environment. Given a user query $q$, an agent attempts to solve the task by producing a trajectory of $\tau = (q, a_1, o_1, a_2, o_2, \dots, a_T, o_T)$, where $a_t$ is the action taken by the agent at step $t$, and $o_t$ is the corresponding environment feedback.
Upon the completion of a trajectory $\tau$, a task verifier evaluates the final outcome and yield a binary reward $r(\tau) \in \{0, 1\}$, indicating the success ($r(\tau) = 1$) or failure ($r(\tau) = 0$) of the trajectory.
At each step, action $a_t$ is generated conditioned on the query and the interaction history observed so far $h_t = (q, a_1, o_1, \dots, a_{t-1}, o_{t-1})$. Throughout this paper, we use the terms \textit{state} and \textit{history} interchangeably to refer to $h_t$.

In this work, we consider the problem of training a capable agent with a \textit{target model} in mind. Specifically, our goal is to curate a trajectory dataset that can most effectively maximize the downstream performance of the target model $\pi_{\text{target}}$ via supervised fine-tuning~\citep{wei2021finetuned,ouyang2022training,chung2024scaling}.

\begin{table*}[t]
\centering
\caption{Comparison of agentic SFT dataset synthesis frameworks based on how the rollout action ($a^{\text{rollout}}_t$) and supervision target ($a^*_t$) are assigned at step $t$. $\pi_{\text{target}}$ and $\pi_{\text{expert}}$ denote the target and expert models, respectively. \ours generates mixed rollouts based on a semantic signal and applies judge-weighted supervision.}
\label{tab:curation_strategies}
\resizebox{\textwidth}{!}{%
\begin{tabular}{@{}llll@{}}
\toprule
\textbf{Data Synthesis Strategy} & \textbf{Rollout Action ($a^{\text{rollout}}_t$)} & \textbf{Supervision Target ($a^*_t$)} & \textbf{Supervised Steps} \\ \midrule
Expert demonstrations & $\pi_{\text{expert}}$ & $\pi_{\text{expert}}$ & All steps \\
Target model rejection sampling & $\pi_{\text{target}}$ & $\pi_{\text{target}}$ & All steps (if trajectory succeeds) \\
Target model rollout w/ expert supervision & $\pi_{\text{target}}$ & $\pi_{\text{expert}}$ & All steps \\
Mixed rollouts (Expert Correction) & $\pi_{\text{target}}$ then $\pi_{\text{expert}}$ & $\pi_{\text{expert}}$ & Expert steps only \\ 
Mixed rollouts (DAgger) & $\pi_{\text{target}}$ or $\pi_{\text{expert}}$ (random heuristic) & $\pi_{\text{expert}}$ & All steps \\ \midrule
\textbf{\ours} & $\pi_{\text{target}}$ (if accepted) / $\pi_{\text{expert}}$ (if rejected) & $\pi_{\text{expert}}$ & Weighted by judge's evaluation \\ \bottomrule
\end{tabular}%
}
\end{table*}

\subsection{Trajectory datasets for SFT}
Supervised fine-tuning (SFT) is a standard practice to train LLM agents~\citep{zeng2024agenttuning,chen-etal-2024-agent,raoof2026openthoughts,blakeman2026nemotron}. 
Given a training dataset $\mathcal{D}$ consists of a collection of history-target action pairs $(h_t, a^*_t)$, the target model $\pi_{\text{target}}$ (parameterized by $\theta$) is trained via SFT by minimizing the objective:
\begin{equation}
\label{eq:sft}
    \mathcal{L}_{\text{SFT}}(\theta) = - \mathbb{E}_{(h_t, a^*_t) \sim \mathcal{D}} \left[ \log \pi_{\theta}(a^*_t \mid h_t) \right]
\end{equation}

Existing works adopt various strategies to construct SFT training datasets for LLM agents~\citep{zeng2024agenttuning,qin2024toolllm,yuan2023scaling,sun2026step,lauffer2025imitation}. We formalize common dataset construction processes through a unified framework by decoupling how the context history $h_t$ is generated from how the supervision signal $a^*_t$ is determined.
Specifically, the context history $h_t$ is formed by rolling out a trajectory with a data generation strategy that takes an action $a^{\text{rollout}}_t$ at each step $t$, where $h_t = (q, a^{\text{rollout}}_1, o_1, \dots, a^{\text{rollout}}_{t-1}, o_{t-1})$.
Independent of the rollout, a supervision target $a^*_t$ can be concurrently determined to serve as the target action given $h_t$. Together, this forms a $(h_t, a^*_t)$ pair for training.

Under this framework, we characterize prior data collection strategies for agentic SFT datasets by how they assign $a^{\text{rollout}}_t$ and $a^*_t$, as summarized in \cref{tab:curation_strategies} and overviewed below:

\noindent \textbf{Expert demonstrations.} The most common approach relies entirely on an expert to generate both the trajectory and the learning signal~\citep{zeng2024agenttuning, qin2024toolllm}. Specifically, both the rollout action and the supervision target are determined by the expert: $a^{\text{rollout}}_t = a^*_t \sim \pi_{\text{expert}}(\cdot \mid h_t)$.
While this yields high-quality data, it strictly limits the context history $h_t$ to the expert's state distribution, leading to a well-known state distribution mismatch when the target model is deployed at inference time~\citep{ross2011reduction,bengio2015scheduled}.

\noindent \textbf{Target model rejection sampling.} To avoid distribution mismatch between the expert and the target model, a straightforward approach is to sample trajectories directly from the target model itself ($a^{\text{rollout}}_t \sim \pi_{\text{target}}(\cdot \mid h_t)$), but only retain the successful ones for supervision ($a^*_t = a^{\text{rollout}}_t \text{ if } r(\tau) = 1$)~\citep{yuan2023scaling}. However, the target model may have low success rates, and the approach restricts the dataset to a narrow subset of already solvable tasks.

\noindent \textbf{Target model rollout with expert supervision.} 
To learn from an expert while staying close to the target model's own state distribution, this approach generates rollouts with the target model ($a^{\text{rollout}}_t \sim \pi_{\text{target}}(\cdot \mid h_t)$), while using the expert to provide supervision signal at every step of the trajectory ($a^*_t \sim \pi_{\text{expert}}(\cdot \mid h_t)$)~\citep{sun2026step}. This follows the spirit of on-policy distillation~\citep{agarwal2024policy}.
However, as the target model dictates the rollout generation, the resulting trajectories may frequently fail, yielding datasets with limited exposures to states near successful completion of the tasks.

\noindent \textbf{Mixed expert-target rollouts.}
These approaches generate rollouts by mixing target model and expert generations to maintain target model's distribution while alleviating the low coverage of states with meaningful task progression.
For instance, On-policy Expert Correction employs an expert to take over rollout generation from a target model mid-trajectory~\citep{lauffer2025imitation}. DAgger-style methods rollout a trajectory by using either the target model or the expert at different steps by chance~\citep{ross2011reduction}.
We include more discussion on related work in \cref{sec:related_work}.


\section{Judge-guided rollout and supervision}
\label{sec:method}

As discussed above, synthesizing optimal target-aware agentic SFT datasets requires balancing two often-conflicting objectives: maintaining the target model's rollout distribution and ensuring broad coverage of trajectories containing critical late-stage states near the successful completion of a task. While prior mixed expert-target rollout approaches attempt to bridge this gap, their mechanisms for intervening with expert actions in a rollout are inherently blind to the target model's actual behavior.

Specifically, On-policy Expert Correction~\citep{lauffer2025imitation} relies on a heuristic intervention point that may trigger prematurely and disrupt a valid trajectory from the target model; or it may trigger too late after unrecoverable errors have occurred.
Furthermore, as the expert completes the entire rollout after the intervention point, the remainder of the trajectory suffers from the same distribution mismatch as pure expert demonstrations.
Similarly, DAgger~\citep{ross2011reduction} employs a random intervention mechanism that overlooks where the target model actually requires assistance. By randomly inserting expert actions, DAgger could overwrite correct target model behaviors, forcing unnecessary distribution shifts; or it could fail to intervene when the target model genuinely makes a mistake.

\begin{figure*}[t]
    \centering
    \includegraphics[width=\textwidth]{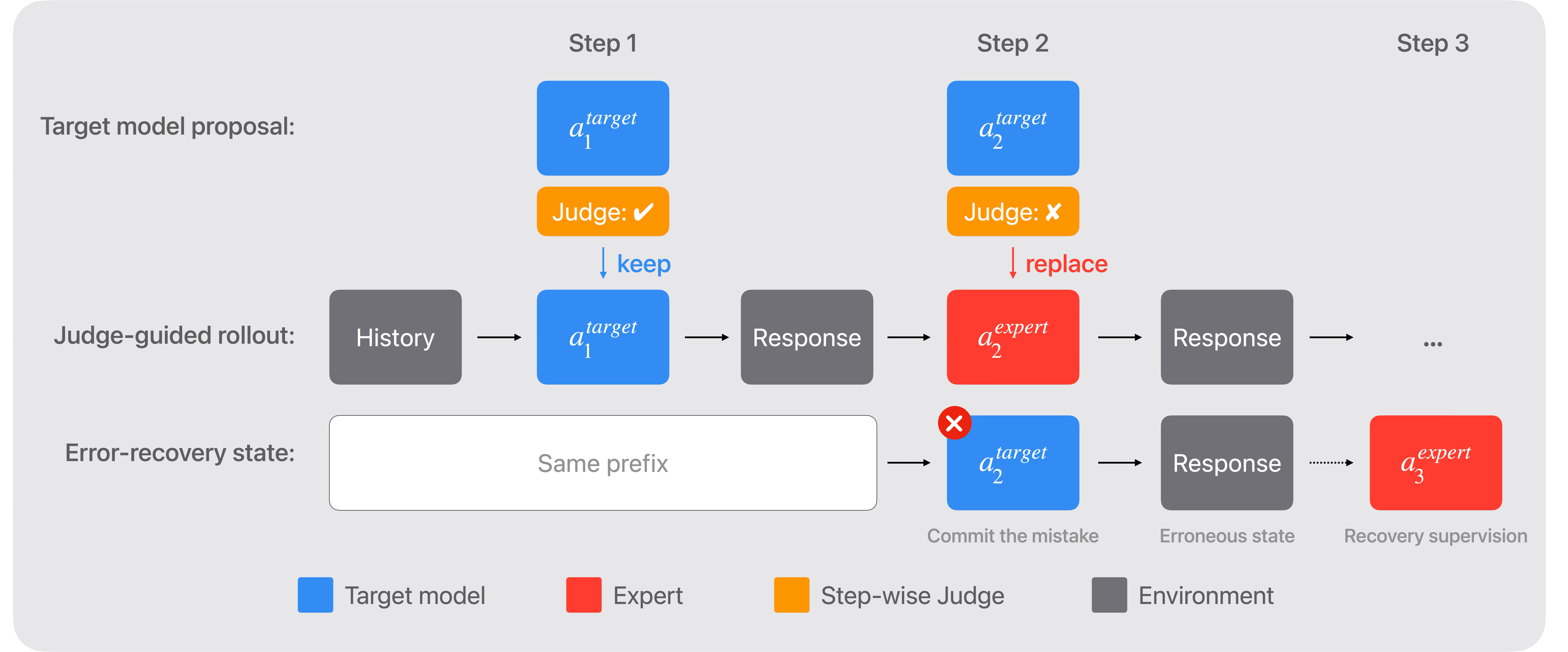}
    \caption{Overview of the \ours framework: Given the history at each step, target model first proposes an action followed by the evaluation of the judge that determines if the action makes reasonable progress of solving the task or not. When accepted by the judge, the target model's action is kept and the rollout continues; when the target model's action is deemed wrong by the judge, expert intervenes and provides a correct action to guide the rollout (\cref{sec:expert_intervention}). At training time, each state in the rollout is paired with the expert's action for supervision, where \ours further uses the judge results to downweight steps where the target model already succeeds (\cref{sec:supervision_reweighting}). Bottom: Additionally, \ours enables collection of erroneous states at steps where the target model fails. These states allow us to obtain corresponding error-recovery supervision.}
    \label{fig:method_overview}
\end{figure*}

To overcome these limitations, we introduce \ours, a framework  that dynamically interleaves the target model and expert during rollout generation based on the \textit{semantic correctness} of the target model's prediction at each step. By employing a judge to provide this per-step evaluation, \ours utilizes the resulting semantic signal to guide trajectory generation with precise expert intervention (\cref{sec:expert_intervention}), and to provide finer-grained control over how expert supervision is applied (\cref{sec:supervision_reweighting}). \cref{fig:method_overview} provides an overview of the framework.

\subsection{Judge-guided expert intervention}
\label{sec:expert_intervention}

In the \ours framework, the target model initiates and drives the rollout for a given task query. At each step $t$ with history $h_t$, the target model generates a candidate action $a^{\text{target}}_t \sim \pi_{\text{target}}(\cdot \mid h_t)$. Rather than immediately committing to this action or blindly overwriting it based on an arbitrary heuristic, we utilize a judge (typically instantiated by the expert) to provide a binary evaluation $\mathcal{J}(a^{\text{target}}_t, h_t) \in \{0, 1\}$. This judge determines whether the proposed action is a semantically correct and productive step toward completing the task: $\mathcal{J}(a^{\text{target}}_t, h_t) = 1$; or otherwise $\mathcal{J}(a^{\text{target}}_t, h_t) = 0$.

Based on the output of $\mathcal{J}$, \ours continues the rollout using the target model's proposed action $a^{\text{target}}_t$ if the judge deems it correct. Conversely, if the proposed step is considered wrong, \ours intervenes in the trajectory with an expert, substituting the action with $a^{\text{expert}}_t \sim \pi_{\text{expert}}(\cdot \mid h_t)$ to keep the trajectory on track. Formally, the rollout action $a^{\text{rollout}}_t$ is assigned conditionally:
\begin{equation}
a^{\text{rollout}}_t = 
\begin{cases} 
a^{\text{target}}_t & \text{if } \mathcal{J}(a^{\text{target}}_t, h_t) = 1\\
a^{\text{expert}}_t & \text{if } \mathcal{J}(a^{\text{target}}_t, h_t) = 0
\end{cases}
\label{eq:rollout_condition}
\end{equation}
We denote the set of histories collected by rolling out the trajectory with these conditionally assigned actions as $\mathcal{H}_{\text{main}}$.
This judge-guided framework provides a principled mechanism for expert intervention, leading to higher trajectory success rates (as seen in \cref{sec:experiments}) while keeping the rollout distribution close to that of the target model.

\paragraph{States for error recovery.}
Beyond keeping the trajectory on track, the per-step judge evaluation in \ours enables precise identification of failure states reached by the target model. This valuable information naturally provides the opportunity to collect states for learning error recovery.
Specifically, when the judge rejects a proposed action ($\mathcal{J}(a^{\text{target}}_t, h_t) = 0$), independent from replacing the step as in \cref{eq:rollout_condition} to continue the main rollout, we can additionally collect an error-inclusive history. By temporarily committing the incorrect action $a^{\text{target}}_t$ to the environment, we advance the trajectory and record the resulting failure state: $h^{\text{error}}_{t+1} = (h_t, a^{\text{target}}_t, o_t)$. This yields a supplementary set of error states, $\mathcal{H}_{\text{error}}$, which we later pair with expert corrective actions to teach error-recovery behaviors, as discussed below.

\subsection{Judge-guided supervision reweighting}
\label{sec:supervision_reweighting}

With the rollout states collected above, we now define the supervision targets for fine-tuning. First, we combine the histories from the main trajectory and the supplementary error states into a unified set: $\mathcal{H}_{\text{all}} = \mathcal{H}_{\text{main}} \cup \mathcal{H}_{\text{error}}$. For every history $h_t \in \mathcal{H}_{\text{all}}$, we query the expert to obtain its action, assigning it as the ground-truth supervision target ($a^*_t \sim \pi_{\text{expert}}(\cdot \mid h_t)$). This yields our final training dataset of history-action pairs:
\begin{equation}
    \mathcal{D} = \{ (h_t, a^*_t) \mid h_t \in \mathcal{H}_{\text{all}} \}
\end{equation}

The standard SFT objective, \cref{eq:sft}, applies a uniform learning signal to every state-action pair in $\mathcal{D}$. This effectively forces the target model to mimic the expert in all situations, regardless of its native competence at each state. However, for a state $h_t$ where the target model already predicts a valid action (i.e., $\mathcal{J}(a^{\text{target}}_t, h_t) = 1$), overwriting this original behavior and forcing unnecessary changes may be sub-optimal.
Conversely, states where the target model fails require strong corrective supervision. Therefore, in \ours, we propose to downweight the learning signal on steps where the target model is already successful---preserving its inherent action---while maintaining full supervision strength for correcting its mistakes.

Formally, for each training pair $(h_t, a^*_t) \in \mathcal{D}$, we assign a supervision weight $\omega_t$ based on the judge's evaluation of the target model's native prediction at that state:
\begin{equation}
\label{eq:supervision_weight}
\omega_t = 
\begin{cases} 
w & \text{if } \mathcal{J}(a^{\text{target}}_t, h_t) = 1 \\
1 & \text{if } \mathcal{J}(a^{\text{target}}_t, h_t) = 0
\end{cases}
\end{equation}
where $w \in [0, 1]$ is a downweighting factor applied to the already-correct states. The target model is then trained using a weighted SFT objective:
\begin{equation}
    \label{eq:reweight_supervision}
    \mathcal{L}_{\text{Weighted-SFT}}(\theta) = - \mathbb{E}_{\mathcal{D}} \left[ \omega_t \log \pi_{\theta}(a^*_t \mid h_t) \right]
\end{equation}
By dynamically reweighting the loss, \cref{eq:reweight_supervision} ensures that the target model is not forced to blindly mimic the expert, allowing it the flexibility to retain its inherent correct behavior. In practice, the value of $w$ can be chosen using a validation set.
\section{Experiments}
\label{sec:experiments}

\subsection{Experimental setup}
\label{sec:experiment_setup}

\noindent \textbf{Benchmarks.} We evaluate \ours on two complex agentic tool-use benchmarks. \textit{AppWorld} tests the agent's ability to act as a personal assistant navigating daily mobile applications with underlying stateful databases~\citep{trivedi2024appworld}. \textit{AutomationBench} evaluates the agent's capability to complete professional workplace workflows by operating specialized enterprise software~\citep{shepard2026automationbench}.
For AppWorld, we utilize the training task split to synthesize datasets for all frameworks considered, and report downstream performances on Test-Normal (Test-N) and Test-Challenge (Test-C) sets. Test-N set includes in-distribution tasks requiring use of seen tools, while Test-C set evaluates the agent's generalization to solving out-of-distribution tasks requiring use of unseen tools.
Following the standard AppWorld evaluation protocol, we report both Task Goal Completion (TGC) and Scenario Goal Completion (SGC) on these sets. TGC measures whether the agent's success rate in successfully completing individual tasks, while SGC is a robustness metric evaluating whether the agent correctly solves all task variants from the same scenario.
For AutomationBench, we follow the same structure and split the original public tasks into Train, Dev, Test-N (in-distribution) and Test-C (out-of-distribution) splits. We include more detail on the benchmark dataset statistics in \cref{sec:benchmark_stats}.

\noindent \textbf{Expert and target models.}
Across all data synthesis frameworks, we employ a highly capable model as the expert. Specifically, we use \texttt{GLM-5.1} for AppWorld and \texttt{GLM-5.2} for AutomationBench~\citep{glm5team2026glm5vibecodingagentic}, as they achieve the highest performance on their respective benchmarks.
We consider \texttt{Qwen3.5-9B} and \texttt{Qwen3.5-2B} as the target models in our experiments~\citep{qwen3.5}.

\noindent \textbf{Baselines.}
We compare \ours against established paradigms for synthesizing agentic SFT datasets, as categorized in \cref{tab:curation_strategies}.

\begin{itemize}[leftmargin=*, nosep]
    \item \textit{Expert Demonstrations:} The standard target model-agnostic approach that samples trajectories from the expert, where the final history-action pairs are constructed from only the successful trajectories with $r(\tau)=1$~\citep{zeng2024agenttuning,qin2024toolllm,yang2026swe}.
    
    \item \textit{Rejection Sampling Finetuning (RFT):} Target model generates the entire trajectory. Successful trajectories are kept to form history-action pairs~\citep{yuan2023scaling}.
    
    \item \textit{Step-level RFT:}  A more granular variant of RFT that additionally removes \textit{incorrect} steps from the RFT trajectories. We use the same judge model as in \ours to evaluate the correctness of each step.
    
    \item \textit{Step-level OPD:} Target model generates the rollouts with expert supervision applied at every step~\citep{sun2026step}. Following the original approach, we do not use the task verifier for trajectory filtration.
    
    \item \textit{On-policy Expert Correction (OEC):} Target model begins the trajectory and the expert takes over at some step $k$ sampled from a uniform distribution $U(0, K)$~\citep{lauffer2025imitation}. History-action pairs are formed only on the expert steps in successful trajectories.
    In the experiments, we set $K=10$ which ensures most trajectories would contain a least one expert step.
    
    \item \textit{DAgger:} In generating a trajectory, each step is generated by the target model with probability $\beta$ or otherwise the expert. Each step in the rollouts is then paired with an expert supervision with no trajectory filtering~\citep{ross2011reduction}.
    Here, we generate the dataset once instead of iteratively with interleaved model training as considered in the original work. We set $\beta=0.5$ in the experiments.
\end{itemize}
For all data generation frameworks including \ours, we report further implementation detail in \cref{sec:implementation_detail}. In addition, we show example trajectories from \ours in \cref{sec:example_trajs}.

\noindent \textbf{Evaluation protocol.}
To compare the overall effectiveness of different data generation strategies, we apply a fixed trajectory generation budget.
In particular, every method generates a fixed number of trajectories (i.e., 8 trajectories per training task) from which the training pairs $(h_t, a^*_t)$ are constructed.
We then apply SFT to the target model with \cref{eq:sft}, and report the downstream pass@1 (averaged over 8 runs) of the fine-tuned model on the benchmarks. We report the detailed statistics of the generated trajectory datasets in \cref{sec:dataset_stats}, and detailed experimental setup including training hyperparameters in \cref{sec:experiment_detail}.

\subsection{Main results}
In \cref{tab:main-results}, we report the downstream performance of the target models fine-tuned using each data synthesis framework on AppWorld and AutomationBench.
Overall, we see \ours demonstrates significant improvements over existing baselines, achieving the highest performance consistently across all target models and benchmarks.

First, compared to Expert Demonstrations, \ours achieves superior performance (up to 10 points on AutomationBench Test-N) by incorporating the target model into the trajectory generation process. We observer similar, although smaller, improvements from other baselines that as well ground trajectory generation with the target model, such as OEC and DAgger.
This validates that while Expert Demonstrations provide the highest-quality trajectories, the inherent distribution mismatch between the expert and the target model makes purely expert data sub-optimal for target model training.

When compared to RFT---a purely target model-driven approach---our results highlight the critical need for expert supervision, particularly when the target model lacks the capability to generate successful rollouts. Interestingly, by utilizing the judge to provide finer-grained feedback at the step level, Step-level RFT improves upon standard RFT. This confirms that the judge's per-step semantic evaluations are indeed valuable. However, Step-level RFT still exhibits a significant performance gap compared to methods that directly leverage the expert for action supervision.

By applying dense expert supervision to target model rollouts, Step-level OPD significantly improves over the RFT variants. For the more capable \texttt{Qwen3.5-9B} model on AppWorld, it ranks among the top-performing baselines, surpassing Expert Demonstrations. This further demonstrates the value of grounding training data close the target model's rollout distribution. However, when applied to the weaker \texttt{Qwen3.5-2B} model or the more complex AutomationBench, Step-level OPD suffers a severe performance drop. This results from that the less capable target model frequently fails on challenging tasks, leading to a rollout distribution with limited coverage of successful, late-stage behaviors.

Finally, while we observe that the mixed-rollout baselines (OEC and DAgger) generally outperform Expert Demonstrations and Step-level OPD---validating the overall benefit of interleaving the expert and target model in trajectory generation, \ours consistently surpasses these strong baselines.
Compared to DAgger, which often serves as the runner-up method, \ours gains an improvement with a margin more than 5 and 7 points on AppWorld and AutomationBench respectively. This is achieved through two key mechanisms. First, it utilizes a judge to provide expert guidance precisely when the target model struggles. Second, it leverages this semantic evaluation to dynamically reweight the supervision strength based on the target model's behavior.

\begin{table*}[t]
    \caption{Downstream performance (pass@1 averaged over 8 runs) of target models fine-tuned using various data synthesis frameworks on AppWorld and AutomationBench. Compared to existing data synthesis strategies, \ours consistently achieves the highest performance across all target model sizes. Note that RFT and Step-level RFT are not applicable on AppWorld with \texttt{Qwen3.5-2B} model given its near-zero untrained model performance.}
    \label{tab:main-results}
    \centering
    \resizebox{\textwidth}{!}{%
    \begin{tabular}{lcccccccccc}
        \toprule
        \multirow{4}{*}{\textbf{Method}}
        & \multicolumn{8}{c}{\textbf{AppWorld}} 
        & \multicolumn{2}{c}{\textbf{AutomationBench}} \\
        \cmidrule(lr){2-9} \cmidrule(lr){10-11}
        & \multicolumn{4}{c}{\texttt{Qwen3.5-2B}}
        & \multicolumn{4}{c}{\texttt{Qwen3.5-9B}} 
        & \multicolumn{2}{c}{\texttt{Qwen3.5-9B}} \\
        \cmidrule(lr){2-5} \cmidrule(lr){6-9} \cmidrule(lr){10-11}
        & \multicolumn{2}{c}{\textbf{Test-N}}
        & \multicolumn{2}{c}{\textbf{Test-C}}
        & \multicolumn{2}{c}{\textbf{Test-N}}
        & \multicolumn{2}{c}{\textbf{Test-C}} 
        & \textbf{Test-N} & \textbf{Test-C} \\
        \cmidrule(lr){2-3}\cmidrule(lr){4-5}\cmidrule(lr){6-7}\cmidrule(lr){8-9}\cmidrule(lr){10-10}\cmidrule(lr){11-11}
        & TGC & SGC & TGC & SGC & TGC & SGC & TGC & SGC & TGC & TGC \\
        \midrule
        {Target Base Model (Zero-shot)}
        & {0.1} & {0.0} & {0.2} & {0.1} 
        & {25.2} & {14.1} & {15.4} & {8.0} 
        & {4.0} & {1.0} \\
        \midrule
        Expert Demonstrations
        & 32.1 & 22.8 & 19.2 & 12.1 
        & 77.4 & 65.9 & 65.0 & 52.7 
        & 17.8 & 8.6 \\
        RFT
        & -- & -- & -- & -- 
        & 52.4 & 38.0 & 37.1 & 24.8 
        & 7.6 & 4.5 \\
        Step-level RFT
        & -- & -- & -- & -- 
        & 59.2 & 45.3 & 46.5 & 32.6 
        & 11.1 & 5.9 \\
        Step-level OPD
        & 17.2 & 10.0 & 8.0 & 4.5
        & 77.8 & 67.6 & 67.2 & 55.0 
        & 13.4 & 8.0 \\
        On-policy Expert Correction
        & 34.8 & 23.9 & 18.1 & 11.4
        & 77.9 & 67.4 & 67.9 & 55.8 
        & 14.0 & 7.9 \\
        DAgger
        & 35.6 & 25.5 & 19.2 & 11.6 
        & 77.0 & 65.2 & 66.0 & 53.1 
        & 21.0 & 15.0 \\
        \textbf{JudgeMix}
        & \textbf{40.9} & \textbf{29.9} & \textbf{23.3} & \textbf{14.6} 
        & \textbf{78.7} & \textbf{67.6} & \textbf{69.8} & \textbf{57.8} 
        & \textbf{28.5} & \textbf{15.1} \\
        \bottomrule
    \end{tabular}%
    }
\end{table*}


\subsection{Ablation studies}

To isolate the contributions of our framework, we conduct ablations on the core mechanisms of \ours.

\paragraph{The value of semantic-based guidance.}
We evaluate the importance of semantic-based expert intervention. Specifically, we investigate whether \ours's performance gains stem from intervening precisely at steps where target model makes a mistake, or merely from the presence of expert actions within the rollout.
To test this, we strip off the weighted supervision component from \ours, where we apply equal supervision weights on all history-action pairs.
We consider DAgger as the comparison baseline where the expert intervene the rollout at random steps regardless of whether the target model succeeds or struggles at those steps.

In \cref{tab:ablation-judge}, we see clear performance improvements moving from DAgger's random intervention mechanism to \ours's semantic-based guidance. The gain is especially clear on scenarios where the initial target model struggles more, for instance, on AppWorld with \texttt{Qwen3.5-2B}.
We conjecture this gain comes from the fact that precise expert guidance can most effectively assist the target model in generating successful trajectories. As seen in \cref{sec:dataset_stats}, judge-guided rollout improves the trajectory pass rate over random intervention by a significant margin of 27 pass@1 points.


\begin{table*}[t]
    \caption{Ablation study on the effect of judge-guided expert intervention. By comparing \ours (without weighted supervision, i.e., $w=1$ in \cref{eq:supervision_weight}) against DAgger, we demonstrate the performance gains of intervening based on semantic correctness rather than random mixing heuristics.}
    \label{tab:ablation-judge}
    \centering
    \resizebox{\linewidth}{!}{%
    \begin{tabular}{lcccccc}
        \toprule
        \multirow{4}{*}{\textbf{Method}}
        & \multicolumn{4}{c}{\textbf{AppWorld}} 
        & \multicolumn{2}{c}{\textbf{AutomationBench}} \\
        \cmidrule(lr){2-5} \cmidrule(lr){6-7}
        & \multicolumn{2}{c}{\texttt{Qwen3.5-2B}}
        & \multicolumn{2}{c}{\texttt{Qwen3.5-9B}} 
        & \multicolumn{2}{c}{\texttt{Qwen3.5-9B}} \\
        \cmidrule(lr){2-3} \cmidrule(lr){4-5} \cmidrule(lr){6-7}
        & \textbf{Test-Normal} & \textbf{Test-Challenge}
        & \textbf{Test-Normal} & \textbf{Test-Challenge} 
        & \textbf{Test-Normal} & \textbf{Test-Challenge} \\
        \cmidrule(lr){2-2}\cmidrule(lr){3-3}\cmidrule(lr){4-4}\cmidrule(lr){5-5}\cmidrule(lr){6-6}\cmidrule(lr){7-7}
        & TGC & TGC & TGC & TGC & TGC & TGC \\
        \midrule
        DAgger
        & 35.6 & 19.2
        & 77.0 & 66.0
        & 21.0 & 15.0 \\
        \ours ($w=1$)
        & \textbf{38.3} & \textbf{21.9}
        & \textbf{77.5} & \textbf{66.2}
        & \textbf{28.5} & \textbf{15.1} \\
        \bottomrule
    \end{tabular}%
    }
\end{table*}

\paragraph{The effect of weighted supervision.} 
To understand the benefits of utilizing the judge's evaluation to dynamically reweight the learning signal on different steps, we show how model's downstream performance changes as $w$ varies in \cref{eq:reweight_supervision}.

\begin{figure}[t]
    \centering
    \includegraphics[width=\linewidth]{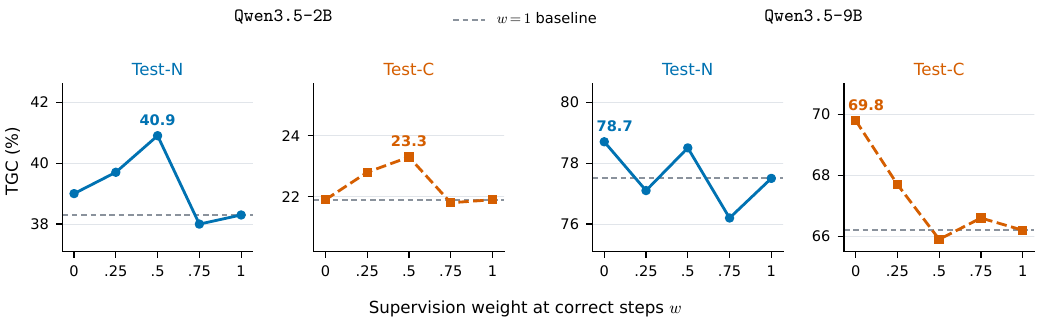}
    \caption{Ablation study on the effect of weighted supervision as $w$ varies in \cref{eq:supervision_weight}. We plot performance on AppWorld Test-N and Test-C with \texttt{Qwen3.5-2B} and \texttt{Qwen3.5-9B} models. Dashed lines mark the $w = 1$ baseline. We observe that down-weighting the supervision signal on the target model's already-successful states improves downstream performance.}
    \label{fig:ablation-selective-supervision}
\end{figure}

As shown in \cref{fig:ablation-selective-supervision}, for \texttt{Qwen3.5-9B} model, decreasing the supervision weight on steps where the target model is already correct consistently improves performance.
In fact, performance peaks at $w=0$, where no supervision is applied at steps where target model generates a correct action.
Interestingly, for the weaker \texttt{Qwen3.5-2B} model, we see a sweet spot at $w=0.5$.
This highlights a trade-off between maximally exploiting expert supervision and preserving the target model's native behavior. We conjecture that even when a weaker model produces a correct action, its underlying reasoning path may be sub-optimal. Consequently, partially overwriting these behaviors remains beneficial, unlike with a stronger target model whose native reasoning is more robust to keep.

\paragraph{The impact of error-recovery data.} 
As discussed in \cref{sec:expert_intervention}, a key advantage of our judge-guided framework is the ability to identify target model errors and collect targeted supervision on how to fix them. This exposes the model to error-correction behaviors if encountered at inference time. Here, we evaluate the performance gains provided by including this error-recovery data.
In \cref{tab:ablation-self-correction}, we see consistent benefits by including error-recovery data in \ours. The improvement is especially significant on \texttt{Qwen3.5-2B} model, with 7.5 and 5 points pass@1 gain on AppWorld Test-N and Test-C sets respectively. We conjecture the larger gain results from that the weaker target model encounters erroneous steps more often, allowing it to benefit more heavily from the error-correction supervision.

\begin{table}[t]
    \caption{Ablation study on the effect of error-recovery data. Including this supplementary data in \ours yields significant improvements, particularly for the weaker \texttt{Qwen3.5-2B} model. This suggests the value of explicitly teaching the target models how to correct from their native failure states.}
    \label{tab:ablation-self-correction}
    \centering
    \resizebox{0.6\linewidth}{!}{%
    \begin{tabular}{lcccc}
        \toprule
        \multirow{4}{*}{\textbf{Method}}
        & \multicolumn{4}{c}{\textbf{AppWorld}} \\
        & \multicolumn{2}{c}{\texttt{Qwen3.5-2B}}
        & \multicolumn{2}{c}{\texttt{Qwen3.5-9B}} \\
        \cmidrule(lr){2-3} \cmidrule(lr){4-5}
        & \textbf{Test-N} & \textbf{Test-C}
        & \textbf{Test-N} & \textbf{Test-C} \\
        \cmidrule(lr){2-2}\cmidrule(lr){3-3}\cmidrule(lr){4-4}\cmidrule(lr){5-5}
        & TGC & TGC & TGC & TGC \\
        \midrule
        JudgeMix w/o error-recovery
        & 33.4 & 18.3
        & 77.5 & 68.2 \\
        JudgeMix
        & \textbf{40.9} & \textbf{23.3}
        & \textbf{78.7} & \textbf{69.8} \\
        \bottomrule
    \end{tabular}%
    }
\end{table}

\subsection{The effect of task verifier}
\label{sec:task_verifier_effect}
So far, we have assumed that each agentic task comes with a ground-truth verifier that provides a trajectory reward $r(\tau)$ indicating the overall success or failure of the rollout. As noted in \cref{sec:experiment_setup}, various baseline approaches as well as \ours utilize this reward to filter for successful trajectories during dataset construction.
However, in many open-ended agentic scenarios, reliable programmatic verifiers are not always available~\citep{qin2024toolllm,lee2026environment,he2024webvoyager,xue2025illusion}. Thus, we investigate how the absence of task-success verifiers impacts the effectiveness of the considered data synthesis frameworks.

In \cref{tab:unfiltered-results}, we see that \ours maintains its strong performance without relying on task verifier for providing trajectory reward, significantly outperforming all baseline strategies on the \texttt{Qwen3.5-2B} model and achieving the highest out-of-distribution generalization (69.1 on Test-C) on the \texttt{Qwen3.5-9B} model.
Notably, methods such as Expert Demonstrations and Step-level RFT suffer greater performance drops when ground-truth verifiers are removed.
Overall, these results demonstrate the robustness of \ours and confirm its applicability to open-ended agentic tasks where final trajectory rewards are not always available.

\begin{table}[t]
    \caption{Downstream performance (pass@1) of target models fine-tuned using various data synthesis frameworks on the \textbf{AppWorld} benchmark, assuming no task verifier for trajectory filtration.}
    \label{tab:unfiltered-results}
    \centering
    \resizebox{0.6\linewidth}{!}{%
    \begin{tabular}{lcccc}
        \toprule
        \multirow{4}{*}{\textbf{Method}}
        & \multicolumn{4}{c}{\textbf{AppWorld}} \\
        \cmidrule(lr){2-5}
        & \multicolumn{2}{c}{\texttt{Qwen3.5-2B}}
        & \multicolumn{2}{c}{\texttt{Qwen3.5-9B}} \\
        \cmidrule(lr){2-3} \cmidrule(lr){4-5}
        & \textbf{Test-N} & \textbf{Test-C}
        & \textbf{Test-N} & \textbf{Test-C} \\
        \cmidrule(lr){2-2}\cmidrule(lr){3-3}\cmidrule(lr){4-4}\cmidrule(lr){5-5}
        & TGC & TGC & TGC & TGC \\
        \midrule
        Expert Demonstrations
        & 32.1 & 17.8
        & 77.9 & 60.7 \\
        Step-level RFT
        & 0.0 & 0.1
        & 49.0 & 36.5 \\
        Step-level OPD
        & 17.2 & 8.0
        & 77.8 & 67.2 \\
        On-policy Expert Correction
        & 33.0 & 18.7
        & \textbf{78.7} & 66.9 \\
        DAgger
        & 35.6 & 19.2
        & 77.0 & 66.0 \\
        JudgeMix
        & \textbf{42.6} & \textbf{24.5}
        & 77.5 & \textbf{69.1} \\
        \bottomrule
    \end{tabular}%
    }
\end{table}

\section{Conclusion}
We introduce \ours, a target-aware data generation framework for agentic SFT datasets. The core design of \ours is to employ a judge to provide step-wise evaluation of the target model's behavior during rollouts. This semantic signal then allows principled expert guidance, collection of error-recovery states, and strategic supervision reweighting.
Empirically, \ours demonstrates consistent performance improvements over the standard model-agnostic approach as well as concurrent model-targeted synthesis frameworks.

\section*{Limitations}
We acknowledge several limitations of the proposed \ours framework.
First, as the framework is intentionally target-aware, the synthesized dataset is coupled to the specific target model used in generation. Unlike standard model-agnostic datasets which can be generated once and used for any downstream models, JudgeMix requires a separate data generation process for each target model to ensure its effectiveness.
We view this as an inherent trade-off between dataset reusability and the optimization of model downstream performance. We leave exploration in potential cross-model transfer to future work.

Second, \ours introduces potential computational overhead during its dataset generation process, as it additionally queries a judge to evaluate the target model's proposed action at every step.
However, with precise expert interventions, the resultant rollouts from \ours achieve higher success rates often with shorter trajectory lengths, making the overall synthesis process highly tractable.

Finally, the effectiveness of JudgeMix is ultimately determined by the reliability of the judge. In open-ended domains where intermediate task progress may be ambiguous, an unreliable judge could produce incorrect evaluation, hurting the rollout quality and the supervision reweighting effectiveness.
However, our empirical results indicate that current open-sourced models are highly capable of providing reliable step-wise evaluations. As seen in our experiments (\cref{tab:main-results}), Step-level RFT yields significant improvements over standard RFT, validating that the judge's evaluation is indeed informative and beneficial in practice.


\bibliographystyle{acl_natbib}
\bibliography{reference}

\clearpage
\appendix
\crefalias{section}{appendix}

\section{Related work}
\label{sec:related_work}

\paragraph{LLM agents and data synthesis.}
LLM agents combine reasoning with actions and environment feedback to solve multi-step tasks, as shown in ReAct \citep{yao2022react}. Training these capabilities requires interaction data beyond conventional instruction-response pairs. ToolLLM \citep{qin2024toolllm} synthesizes tool-use demonstrations while AgentTuning \citep{zeng2024agenttuning} and AgentBank \citep{song-etal-2024-agentbank} curate interaction trajectories across diverse agent tasks. Subsequent works expand the scale, diversity, and reliability of synthetic supervision: AgentGen \citep{hu2025agentgen} generates environments and planning tasks; APIGen-MT \citep{prabhakar2026apigen} combines task blueprints with simulated human-agent interactions and SWE-smith \citep{yang2026swe} constructs executable software engineering tasks at scale. More recently, OpenThoughts-Agent \citep{raoof2026openthoughts} systematically studies data curation for broadly capable agents, while Simulate to Generalize \citep{lee2026environment} uses LLM world models to synthesize stateful API interaction trajectories without requiring executable environments. Moving beyond fixed datasets, CoEvolve \citep{yang2026coevolve} uses rollout-derived weakness signals to guide task synthesis as the agent learns. Our approach complements these approaches with target-aware data synthesis, using step-level semantic judgements to guide expert intervention and supervision weighting.


\paragraph{Target-aware agentic data curation with expert guidance.}

While most existing agentic data curation methods rely primarily on strong expert models to generate training trajectories, recent approaches have explored target model-aware data curation, where expert models provide additional guidance to improve the quality of on-policy trajectories generated by the target model.
Specifically, \citet{lauffer2025imitation} and \citet{li2026policy} enable expert models to participate in the rollout generation process through manually designed curricula,
while \citet{sun2026step} leverages expert models to provide dense supervision signals at each turn of a rollout.
Alternatively, \citet{da2025agent} employs trajectory-level expert feedback to augment rollouts generated by the target model.
In contrast, our proposed JudgeMix similarly incorporates expert models into rollout generation while simultaneously providing turn-level guidance through an LLM-as-a-judge framework.
\section{Benchmark dataset statistics}
\label{sec:benchmark_stats}

We include detailed benchmark dataset statistics on AppWorld and AutomationBench in \cref{tab:appworld-splits} and \cref{tab:automationbench-splits} respectively.

As AutomationBench does not come with training splits, we construct our own custom splits over its 600 public tasks, which cover six business domains with 100 tasks each. In order to construct an out-of-distribution test set, we hold out the entire Support domain as Test-Challenge, where all tasks under this domain utilizes apps that never appear in other domains.

\section{Implementation detail}
\label{sec:implementation_detail}
We report the implementation detail for all data generation frameworks considered in this work. For generating trajectories, we utilize vLLM~\citep{kwon2023efficient} for model serving and rely on the agentic harness provided by the respective benchmarks, AppWorld~\citep{trivedi2024appworld} and AutomationBench~\citep{shepard2026automationbench}, to specify the task prompts and the interaction interface with the environments. We refer readers to the original work for detailed information.
For both AppWorld and AutomationBench, each task comes with a verifier that outputs $r(\tau)=\{0, 1\}$, indicating whether the trajectory successfully solves the corresponding task or not. Methods including Expert Demonstrations, RFT, Step-level RFT, OEC, and \ours utilize this trajectory annotation to keep only the successful trajectories for constructing the training dataset $\mathcal{D}$. We refer readers to \cref{sec:task_verifier_effect} for results without access to the task verifiers. Below, we include the implementation detail specific to different methods.

\paragraph{JudgeMix.}
For \ours, in generating a step from either the expert or the target model, we utilize the same provided benchmark harnesses as with other data generation methods.
For step-wise judge evaluation, we instantiate the judge with the expert, using \texttt{GLM-5.1} for AppWorld and \texttt{GLM-5.2} for AutomationBench. We include the prompts for the judge in \cref{prompt:judge-appworld} (AppWorld) and \cref{prompt:judge-automationbench} (AutomationBench).


\begin{table}[t]
  \centering
  \begin{minipage}[t]{0.48\linewidth}
  \centering
  \small
  \caption{Number of tasks in AppWorld dataset splits. Each scenario comes with three task variants. Test-Challenge tasks each use at least one app that never appears in Train or Dev. This split measures model's generalization to unseen apps.}
  \label{tab:appworld-splits}
  \begin{tabular}{lrr}
    \toprule
    Split & Scenarios & Tasks \\
    \midrule
    Train          & 30  & 90  \\
    Dev            & 19  & 57  \\
    Test-Normal    & 56  & 168 \\
    Test-Challenge & 139 & 417 \\
    \midrule
    Total          & 244 & 732 \\
    \bottomrule
  \end{tabular}
  
  \end{minipage}%
  \hfill
  \begin{minipage}[t]{0.48\linewidth}
    
  \centering
  \small
  \caption{Number of tasks in AutomationBench custom dataset splits. We split the original 600 public tasks into Train, Dev, Test-Normal (Test-N), and Test-Challenge (Test-C). Dev and Test-Normal are disjoint task sets drawn from the same five domains covered by Train. Test-Challenge consists of all tasks in the Support domain, which never appears in Train, and tests generalization to unseen apps and workflows.}
  \label{tab:automationbench-splits}
  \begin{tabular}{lrrrr}
    \toprule
    Domain & Train & Dev & Test-N & Test-C \\
    \midrule
    Sales      & 70  & 10  & 20  & --- \\
    Marketing  & 70  & 10  & 20  & --- \\
    Operations & 70  & 10  & 20  & --- \\
    Finance    & 70  & 10  & 20  & --- \\
    HR         & 70  & 10  & 20  & --- \\
    Support    & --- & --- & --- & 100 \\
    \midrule
    Total      & 350 & 50  & 100 & 100 \\
    \bottomrule
  \end{tabular}
    
  \end{minipage}
\end{table}

\section{Example trajectories}
\label{sec:example_trajs}
We show example \ours trajectories generated on AppWorld in \cref{fig:appworld_example} (single-app task) and \cref{fig:appworld_example_1} (cross-app task).
Although all users and data in AppWorld are synthetic, we replace app names and personal
details with placeholder tags: app names become generic tags (e.g., \texttt{<music\_app>},
\texttt{<payment\_app>}), and user names, email addresses, phone numbers, passwords, access
tokens, and other fictitious entities such as artist and person names are replaced in the
same way (e.g., \texttt{<email\_address>}, \texttt{<artist\_1>}).
Note that the example tasks and data are synthetically generated and do not correspond to real users and data.

The main rollout follows the target model's generation when the proposed action is accepted by the judge (green colored steps), while the rollout uses expert's generation (blue colored steps) at steps where the target model's action is rejected. These rejected steps (red colored) are used to collect error-recovery supervision.
Note that the trajectory illustrates only the states in \ours. For each state $h_t$, we collect the corresponding expert's action as the supervision target ($a^*_t \sim \pi_\text{expert}(\cdot \mid h_t)$), as described in \cref{sec:supervision_reweighting}.

\section{Trajectory dataset statistics}
\label{sec:dataset_stats}

In \cref{tab:dataset-statistics}, we include the statistics of the training datasets generated from different frameworks. Note that across different frameworks, we always fix the total number of generated trajectories.
We see that \ours has a much higher pass@1 than DAgger, suggesting that expert intervention based on judge's semantic evaluation is more effective in keeping the rollout on track than random intervention.
In addition, compared to methods relying purely on target model for rollouts, \ours has much shorter trajectory length.

\begin{table*}[t]
    \caption{Dataset statistics for the training trajectories synthesized by different data generation frameworks. We report the total number of trajectories, the trajectory success rate (pass@1) prior to any filtration, and the average number of steps per trajectory.}
    \label{tab:dataset-statistics}
    \centering
    \resizebox{\textwidth}{!}{%
    \begin{tabular}{lccccccccc}
        \toprule
        \multirow{3}{*}{\textbf{Method}}
        & \multicolumn{6}{c}{\textbf{AppWorld}}
        & \multicolumn{3}{c}{\textbf{AutomationBench}} \\
        \cmidrule(lr){2-7} \cmidrule(lr){8-10}
        & \multicolumn{3}{c}{\texttt{Qwen3.5-2B}}
        & \multicolumn{3}{c}{\texttt{Qwen3.5-9B}}
        & \multicolumn{3}{c}{\texttt{Qwen3.5-9B}} \\
        \cmidrule(lr){2-4} \cmidrule(lr){5-7} \cmidrule(lr){8-10}
        & Total & Traj. & Avg.
        & Total & Traj. & Avg.
        & Total & Traj. & Avg. \\
        & Trajs & pass@1 & Steps
        & Trajs & pass@1 & Steps
        & Trajs & pass@1 & Steps \\
        \midrule
        Expert Demonstrations
        & 720 & 89.7 & 15.0
        & 720 & 89.7 & 15.0
        & 3,200 & 19.7 & 27.5 \\
        Step-level RFT
        & 720 & 0.0 & 31.6
        & 720 & 32.8 & 18.4
        & 3,200 & 4.0 & 36.8 \\
        Step-level OPD
        & 720 & 0.0 & 31.6
        & 720 & 32.8 & 18.4
        & 3,200 & 4.0 & 36.8 \\
        On-policy Expert Correction
        & 720 & 75.7 & 17.5
        & 720 & 81.1 & 15.4
        & 3,200 & 18.7 & 30.0 \\
        DAgger
        & 720 & 51.2 & 18.4
        & 720 & 60.2 & 15.3
        & 3,200 & 17.3 & 25.4 \\
        \textbf{JudgeMix}
        & 720 & 78.5 & 14.7
        & 720 & 72.1 & 15.3
        & 3,200 & 22.3 & 26.6 \\
        \bottomrule
    \end{tabular}%
    }
\end{table*}

\section{Experimental detail}
\label{sec:experiment_detail}

\paragraph{Training setup.} For all SFT experiments, we conduct full-model finetuning with AdamW optimizer with an LR schedule of linear warmup then decay to 0. Warmup is always set to 1\% of the total training steps. We also use fixed weight decay of 0.01, gradient clipping with max norm 1.0. We set a max example context length of 32,768 tokens for AppWorld, and 65,536 tokens for AutomationBench. Examples exceeding the max length is dropped from the training set. We train with batch size 1 (with sequence packing up to corresponding max example length) per GPU on a single node with 8 NVIDIA B200 GPUs. 
On AppWorld, we set learning rate to 1e-5 and train for 10 epochs, which we find to yield most stable training.
On AutomationBench, we set learning rate to 2e-5 and train for 10 epochs.
Additionally, for \ours, we select $w$ in \cref{eq:supervision_weight} from a grid search in $\{0, 0.25, 0.5, 0.75, 1.0\}$ using the Dev set.

\paragraph{Evaluation setup.}
Given a trained model checkpoint, we evaluate the model using the released harness from respective benchmarks. For all pass@1 numbers reported, we average over 8 independent evaluation runs.

\clearpage
\onecolumn

\promptbox[darkgray]{Judge system prompt for AppWorld}{prompts/appworld_system.txt}
\promptfigure[darkgray]{Judge user prompt template for AppWorld}{prompts/appworld_user.txt}
  {System and user prompts used for step-wise judge on AppWorld.}{prompt:judge-appworld}

\newpage
\promptbox[darkgray]{Judge system prompt for AutomationBench}{prompts/automationbench_system.txt}
\promptfigure[darkgray]{Judge user prompt template for AutomationBench}{prompts/automationbench_user.txt}
  {System and user prompts used for step-wise judge on AutomationBench.}{prompt:judge-automationbench}

\newpage
\begin{center}
\definecolor{trajaccept}{HTML}{2E7D32}
\definecolor{trajreject}{HTML}{C62828}
\definecolor{trajexpert}{HTML}{1F5FA8}
\definecolor{trajbranch}{HTML}{E07B00}
\definecolor{trajtask}{HTML}{555555}
\lstdefinestyle{trajcode}{basicstyle=\ttfamily\scriptsize, breaklines=true, columns=fullflexible,
  keepspaces=true, showstringspaces=false, upquote=true, backgroundcolor=\color{black!4},
  framesep=3pt, frame=single, rulecolor=\color{black!12}, aboveskip=2pt, belowskip=4pt,
  xleftmargin=3pt, xrightmargin=3pt}
\lstdefinestyle{trajout}{style=trajcode, basicstyle=\ttfamily\tiny, backgroundcolor=\color{black!8},
  rulecolor=\color{black!8}}
\tcbset{trajstep/.style={enhanced, breakable, boxrule=0.6pt, arc=2pt, left=5pt, right=5pt, top=3pt,
  bottom=3pt, colback=#1!3, colframe=#1, coltitle=white, fonttitle=\bfseries\sffamily\footnotesize,
  fontupper=\footnotesize, before skip=5pt, after skip=5pt}}
\newcommand{\trajpart}[1]{\par\vspace{1pt}{\sffamily\scriptsize\bfseries\color{black!65}#1}\par}
\newcommand{\trajjudge}[3]{%
  \par\vspace{2pt}\noindent\colorbox{#1!8}{\parbox{\dimexpr\linewidth-2\fboxsep}{\scriptsize
  {\sffamily\bfseries\color{#1}Judge: #2}\quad #3}}\par\vspace{3pt}}

\begin{tcolorbox}[trajstep=trajtask, title={Example AppWorld single-app task (success)}]
Name the artist most recommended to me on \textless{}music\_app\textgreater{}.
\end{tcolorbox}
\begin{tcolorbox}[trajstep=trajaccept, title={Step 1 \textperiodcentered{} Target Model (accepted)}]
\trajpart{Target Model response}
\begin{lstlisting}[style=trajcode]
print(apis.api_docs.show_app_descriptions())
\end{lstlisting}
\trajjudge{trajaccept}{accept $\checkmark$}{Listing the available apps is a sensible first step; no API is called without its documentation.}
\trajpart{Output}
\begin{lstlisting}[style=trajout]
[
 {
  "name": "api_docs",
  "description": "An app to search and explore API documentation."
 },
 {
  "name": "supervisor",
  "description": "An app to access supervisor's personal information, account credentials, addresses, payment cards, and manage the assigned task."
[... 30 more lines ...]
  "name": "<note_app>",
  "description": "A note-taking app to create and manage notes."
 },
 {
  "name": "<todo_app>",
  "description": "A task management app to manage todo lists and collaborate on them with others."
 }
]
\end{lstlisting}
\end{tcolorbox}
\begin{tcolorbox}[trajstep=trajaccept, title={Step 2 \textperiodcentered{} Target Model (accepted)}]
\trajpart{Target Model response}
\begin{lstlisting}[style=trajcode]
print(apis.api_docs.show_api_descriptions(app_name='<music_app>'))
\end{lstlisting}
\trajjudge{trajaccept}{accept $\checkmark$}{Looking up the \textless{}music\_app\textgreater{} APIs is the logical next step before calling any of them.}
\trajpart{Output}
\begin{lstlisting}[style=trajout]
[
 {
  "name": "show_account",
  "description": "Show your account information. Unlike show_profile, this includes private information."
 },
 {
  "name": "signup",
  "description": "Sign up to create account."
[... 354 more lines ...]
  "name": "show_premium_subscriptions",
  "description": "Show your premium subscription history."
 },
 {
  "name": "download_premium_subscription_receipt",
  "description": "Download the receipt for a premium subscription."
 }
]
\end{lstlisting}
\end{tcolorbox}
\begin{tcolorbox}[trajstep=trajaccept, title={Step 3 \textperiodcentered{} Target Model (accepted)}]
\trajpart{Target Model response}
\begin{lstlisting}[style=trajcode]
# First, let me get the account credentials for <music_app> login
print(apis.supervisor.show_account_passwords())
\end{lstlisting}
\trajjudge{trajaccept}{accept $\checkmark$}{Borderline: \texttt{show\_account\_passwords} is called without reading its documentation, but the task prompt names this API for getting credentials, so the step is accepted.}
\trajpart{Output}
\begin{lstlisting}[style=trajout]
[
 {
  "account_name": "<shopping_app>",
  "password": "<password>"
 },
 {
  "account_name": "<file_app>",
  "password": "<password>"
[... 22 more lines ...]
  "account_name": "<todo_app>",
  "password": "<password>"
 },
 {
  "account_name": "<payment_app>",
  "password": "<password>"
 }
]
\end{lstlisting}
\end{tcolorbox}
\begin{tcbraster}[raster columns=2, raster equal height=rows, raster column skip=6pt,
  raster before skip=5pt, raster after skip=5pt]
\begin{tcolorbox}[trajstep=trajreject, unbreakable, title={Step 4 \textperiodcentered{} Target Model (rejected): States for error-recovery}]
\trajpart{Target Model response}
\begin{lstlisting}[style=trajcode]
# Login to <music_app> first
login_response = apis.<music_app>.login(email='<email_address>', password='<password>')
print("Login response:", login_response)

# Get personalized song recommendations
recommendations = apis.<music_app>.show_recommendations()
print("Recommendations:", recommendations)
\end{lstlisting}
\trajjudge{trajreject}{reject $\times$}{\texttt{login} and \texttt{show\_recommendations} are called without reading their documentation (criterion 1), so the parameter names are unverified (e.g. \texttt{username} vs. \texttt{email}).}
\trajpart{Output}
\begin{lstlisting}[style=trajout]
Execution failed. Traceback:
  File "<python-input>", line 2, in <module>
    login_response = apis.<music_app>.login(email='<email_address>', password='<password>')
Exception: Unexpected parameter 'email' passed to the login API of the <music_app> app. Allowed parameters are: ['username', 'password']
\end{lstlisting}
\end{tcolorbox}
\begin{tcolorbox}[trajstep=trajexpert, unbreakable, title={Step 4 \textperiodcentered{} Expert replacement}]
\trajpart{Expert response}
\begin{lstlisting}[style=trajcode]
# Login to <music_app>
print(apis.<music_app>.login(email='<email_address>', password='<password>'))
\end{lstlisting}
\trajpart{Output}
\begin{lstlisting}[style=trajout]
Execution failed. Traceback:
  File "<python-input>", line 2, in <module>
    print(apis.<music_app>.login(email='<email_address>', password='<password>'))
Exception: Unexpected parameter 'email' passed to the login API of the <music_app> app. Allowed parameters are: ['username', 'password']
\end{lstlisting}
\end{tcolorbox}
\end{tcbraster}
\begin{tcbraster}[raster columns=2, raster equal height=rows, raster column skip=6pt,
  raster before skip=5pt, raster after skip=5pt]
\begin{tcolorbox}[trajstep=trajreject, unbreakable, title={Step 5 \textperiodcentered{} Target Model (rejected): States for error-recovery}]
\trajpart{Target Model response}
\begin{lstlisting}[style=trajcode]
print(apis.<music_app>.show_account(user_id='<email_address>'))
\end{lstlisting}
\trajjudge{trajreject}{reject $\times$}{\texttt{show\_account} is called before logging in and without its documentation; the step 4 error already said \texttt{login} takes \texttt{username}, so the login should be fixed first.}
\trajpart{Output}
\begin{lstlisting}[style=trajout]
Execution failed. Traceback:
  File "<python-input>", line 1, in <module>
    print(apis.<music_app>.show_account(user_id='<email_address>'))
Exception: Unexpected parameter 'user_id' passed to the show_account API of the <music_app> app. Allowed parameters are: ['access_token']
\end{lstlisting}
\end{tcolorbox}
\begin{tcolorbox}[trajstep=trajexpert, unbreakable, title={Step 5 \textperiodcentered{} Expert replacement}]
\trajpart{Expert response}
\begin{lstlisting}[style=trajcode]
# Login to <music_app> with the correct parameter name
print(apis.<music_app>.login(username='<email_address>', password='<password>'))
\end{lstlisting}
\trajpart{Output}
\begin{lstlisting}[style=trajout]
{
 "access_token": "<access_token>",
 "token_type": "Bearer"
}
\end{lstlisting}
\end{tcolorbox}
\end{tcbraster}
\begin{tcolorbox}[trajstep=trajaccept, title={Step 6 \textperiodcentered{} Target Model (accepted)}]
\trajpart{Target Model response}
\begin{lstlisting}[style=trajcode]
# Get the specification for show_recommendations API
print(apis.api_docs.show_api_doc(app_name='<music_app>', api_name='show_recommendations'))
\end{lstlisting}
\trajjudge{trajaccept}{accept $\checkmark$}{Reads the \texttt{show\_recommendations} documentation before calling it.}
\trajpart{Output}
\begin{lstlisting}[style=trajout]
{
 "app_name": "<music_app>",
 "api_name": "show_recommendations",
 "path": "/<music_app>/recommendations",
 "method": "GET",
 "description": "Get personalized song recommendations for the user.",
 "parameters": [
  {
[... 41 more lines ...]
    ]
   }
  ],
  "failure": {
   "message": "string"
  }
 }
}
\end{lstlisting}
\end{tcolorbox}
\begin{tcolorbox}[trajstep=trajaccept, title={Step 7 \textperiodcentered{} Target Model (accepted)}]
\trajpart{Target Model response}
\begin{lstlisting}[style=trajcode]
# Call show_recommendations API to get personalized song recommendations
# Requesting first page with default limit to see the recommendations
recommendations = apis.<music_app>.show_recommendations(access_token='<access_token>', page_index=0, page_limit=20)

# Store the first page results to check for pagination
print(recommendations)
\end{lstlisting}
\trajjudge{trajaccept}{accept $\checkmark$}{Calls \texttt{show\_recommendations} with documented parameters and the token from step 5; fetching only the first page is accepted as a first look at the data.}
\trajpart{Output}
\begin{lstlisting}[style=trajout]
[
 {
  "song_id": 9,
  "title": "When Fate Becomes a Foe",
  "album_id": 2,
  "album_title": "Celestial Harmonies",
  "duration": 196,
  "artists": [
[... 250 more lines ...]
  "artists": [
   {
    "id": 35,
    "name": "<artist_13>"
   }
  ]
 }
]
\end{lstlisting}
\end{tcolorbox}
\begin{tcolorbox}[trajstep=trajaccept, title={Step 8 \textperiodcentered{} Target Model (accepted)}]
\trajpart{Target Model response}
\begin{lstlisting}[style=trajcode]
# Check for more pages of recommendations
recommendations_page1 = apis.<music_app>.show_recommendations(access_token='<access_token>', page_index=1, page_limit=20)

recommendations_page2 = apis.<music_app>.show_recommendations(access_token='<access_token>', page_index=2, page_limit=20)

recommendations_page3 = apis.<music_app>.show_recommendations(access_token='<access_token>', page_index=3, page_limit=20)

# Store recommendations on each page for analysis
page0_rec = recommendations
page1_rec = recommendations_page1
page2_rec = recommendations_page2
page3_rec = recommendations_page3

# Now I need to analyze which artist appears most frequently across all pages
def count_artist_occurrences(recs):
    artist_counts = {}
    for song in recs:
        for artist in song['artists']:
            artist_counts[artist['name']] = artist_counts.get(artist['name'], 0) + 1
    return artist_counts

all_artist_counts = {}
for page_rec in [page0_rec, page1_rec, page2_rec, page3_rec]:
    counts = count_artist_occurrences(page_rec)
    for artist, count in counts.items():
        all_artist_counts[artist] = all_artist_counts.get(artist, 0) + count

# Find the most recommended artist
most_recommended_artist = max(all_artist_counts, key=all_artist_counts.get)
print(f"API Response on page 1: {len(page0_rec)} songs")
print(f"API Response on page 2: {len(page1_rec) if page1_rec else 0} songs")
print(f"API Response on page 3: {len(page2_rec) if page2_rec else 0} songs")
print(f"API Response on page 4: {len(page3_rec) if page3_rec else 0} songs")
print(f"Artist occurrence counts: {all_artist_counts}")
print(f"Most recommended artist: {most_recommended_artist}")

# If pages are empty, assume we only have one page
if page1_rec == [] and page2_rec == [] and page3_rec == []:
    most_recommended_artist = max(count_artist_occurrences(page0_rec).items(), key=lambda x: x[1])[0]
    print(f"Most recommended artist (from single page): {most_recommended_artist}")
else:
    print(f"Most recommended artist (across all pages): {most_recommended_artist}")
\end{lstlisting}
\trajjudge{trajaccept}{accept $\checkmark$}{Fetches the remaining pages and counts artist occurrences across all of them; some redundancy in the code, but the logic is correct.}
\trajpart{Output}
\begin{lstlisting}[style=trajout]
API Response on page 1: 20 songs
API Response on page 2: 1 songs
API Response on page 3: 0 songs
API Response on page 4: 0 songs
Artist occurrence counts: {'<artist_0>': 2, '<artist_1>': 3, '<artist_2>': 2, '<artist_3>': 1, '<artist_4>': 1, '<artist_5>': 1, '<artist_6>': 2, '<artist_7>': 1, '<artist_8>': 1, '<artist_9>': 2, '<artist_10>': 2, '<artist_11>': 1, '<artist_12>': 1, '<artist_13>': 1, 'Noah Bennett': 1}
Most recommended artist: <artist_1>
Most recommended artist (across all pages): <artist_1>
\end{lstlisting}
\end{tcolorbox}
\begin{tcolorbox}[trajstep=trajaccept, title={Step 9 \textperiodcentered{} Target Model (accepted)}]
\trajpart{Target Model response}
\begin{lstlisting}[style=trajcode]
# Complete the task with the answer
apis.supervisor.complete_task(answer="<artist_1>")
\end{lstlisting}
\trajjudge{trajaccept}{accept $\checkmark$}{All 21 recommendations were counted and \textless{}artist\_1\textgreater{} appears most often (3 times), so completing the task with this answer is correct.}
\trajpart{Output}
\begin{lstlisting}[style=trajout]
Execution successful.
\end{lstlisting}
\end{tcolorbox}

\captionof{figure}{Example \ours trajectory on AppWorld with \texttt{Qwen3.5-9B} target model and \texttt{GLM-5.1} as the expert and judge. The rollout follows target model's generation when the proposed action is accepted by the judge (\textcolor{trajaccept}{green colored steps}), while using expert's generation (\textcolor{trajexpert}{blue colored steps}) when the proposed action is rejected. The rejected steps (\textcolor{trajreject}{red colored}) are used to collect error-recovery supervision.}
\label{fig:appworld_example}
\end{center}

\newpage
\begin{center}
\definecolor{trajaccept}{HTML}{2E7D32}
\definecolor{trajreject}{HTML}{C62828}
\definecolor{trajexpert}{HTML}{1F5FA8}
\definecolor{trajbranch}{HTML}{E07B00}
\definecolor{trajtask}{HTML}{555555}
\lstdefinestyle{trajcode}{basicstyle=\ttfamily\scriptsize, breaklines=true, columns=fullflexible,
  keepspaces=true, showstringspaces=false, upquote=true, backgroundcolor=\color{black!4},
  framesep=3pt, frame=single, rulecolor=\color{black!12}, aboveskip=2pt, belowskip=4pt,
  xleftmargin=3pt, xrightmargin=3pt}
\lstdefinestyle{trajout}{style=trajcode, basicstyle=\ttfamily\tiny, backgroundcolor=\color{black!8},
  rulecolor=\color{black!8}}
\tcbset{trajstep/.style={enhanced, breakable, boxrule=0.6pt, arc=2pt, left=5pt, right=5pt, top=3pt,
  bottom=3pt, colback=#1!3, colframe=#1, coltitle=white, fonttitle=\bfseries\sffamily\footnotesize,
  fontupper=\footnotesize, before skip=5pt, after skip=5pt}}
\newcommand{\trajpart}[1]{\par\vspace{1pt}{\sffamily\scriptsize\bfseries\color{black!65}#1}\par}
\newcommand{\trajjudge}[3]{%
  \par\vspace{2pt}\noindent\colorbox{#1!8}{\parbox{\dimexpr\linewidth-2\fboxsep}{\scriptsize
  {\sffamily\bfseries\color{#1}Judge: #2}\quad #3}}\par\vspace{3pt}}

\begin{tcolorbox}[trajstep=trajtask, title={Example AppWorld cross-app task (success)}]
Like all the \textless{}payment\_app\textgreater{} transactions from yesterday involving any of my siblings on my \textless{}payment\_app\textgreater{} social feed.
\end{tcolorbox}
\begin{tcbraster}[raster columns=2, raster equal height=rows, raster column skip=6pt,
  raster before skip=5pt, raster after skip=5pt]
\begin{tcolorbox}[trajstep=trajreject, unbreakable, title={Step 1 \textperiodcentered{} Target Model (rejected): States for error-recovery}]
\trajpart{Target Model response}
\begin{lstlisting}[style=trajcode]
```python
{"message": "I'll help you like all <payment_app> transactions from yesterday involving your siblings. Let me first explore what APIs are available and understand the system.", "code": "print(apis.api_docs.show_app_descriptions())"}
```
</think>

I need to systematically explore the available apps and APIs to complete this task. Let me start by getting the list of available apps.

```python
{"message": "Let me explore the available apps to find <payment_app> and other social platforms.", "code": "print(apis.api_docs.show_app_descriptions())"}
```
</think>
\end{lstlisting}
\trajjudge{trajreject}{reject $\times$}{The code block is a JSON-like dictionary with \texttt{message} and \texttt{code} keys rather than Python, so the API call would never run.}
\trajpart{Output}
\begin{lstlisting}[style=trajout]
Execution successful.
\end{lstlisting}
\end{tcolorbox}
\begin{tcolorbox}[trajstep=trajexpert, unbreakable, title={Step 1 \textperiodcentered{} Expert replacement}]
\trajpart{Expert response}
\begin{lstlisting}[style=trajcode]
print(apis.api_docs.show_app_descriptions())
\end{lstlisting}
\trajpart{Output}
\begin{lstlisting}[style=trajout]
[
 {
  "name": "api_docs",
  "description": "An app to search and explore API documentation."
 },
 {
  "name": "supervisor",
  "description": "An app to access supervisor's personal information, account credentials, addresses, payment cards, and manage the assigned task."
[... 30 more lines ...]
  "name": "<note_app>",
  "description": "A note-taking app to create and manage notes."
 },
 {
  "name": "<todo_app>",
  "description": "A task management app to manage todo lists and collaborate on them with others."
 }
]
\end{lstlisting}
\end{tcolorbox}
\end{tcbraster}
\begin{tcolorbox}[trajstep=trajaccept, title={Step 2 \textperiodcentered{} Target Model (accepted)}]
\trajpart{Target Model response}
\begin{lstlisting}[style=trajcode]
print(apis.api_docs.show_api_descriptions(app_name='<payment_app>'))
\end{lstlisting}
\trajjudge{trajaccept}{accept $\checkmark$}{Listing the \textless{}payment\_app\textgreater{} APIs is the logical next step before calling any of them.}
\trajpart{Output}
\begin{lstlisting}[style=trajout]
[
 {
  "name": "show_account",
  "description": "Show your account information. Unlike show_profile, this includes private information."
 },
 {
  "name": "signup",
  "description": "Sign up to create account."
[... 202 more lines ...]
  "name": "delete_notification",
  "description": "Delete a notification."
 },
 {
  "name": "mark_notification",
  "description": "Mark a notification as read or unread."
 }
]
\end{lstlisting}
\end{tcolorbox}
\begin{tcbraster}[raster columns=2, raster equal height=rows, raster column skip=6pt,
  raster before skip=5pt, raster after skip=5pt]
\begin{tcolorbox}[trajstep=trajreject, unbreakable, title={Step 3 \textperiodcentered{} Target Model (rejected): States for error-recovery}]
\trajpart{Target Model response}
\begin{lstlisting}[style=trajcode]
# Get today's date to calculate yesterday's date
from datetime import datetime, timedelta
yesterday = (datetime.now() - timedelta(days=1)).strftime('%Y-%m-%d')
print(f"Today's date: {datetime.now().strftime('%Y-%m-%d')}")
print(f"Yesterday's date: {yesterday}")

# Get account credentials for <payment_app> login
print(apis.supervisor.show_account_passwords())

# Get siblings from supervisor (family information)
print(apis.supervisor.show_account())
\end{lstlisting}
\trajjudge{trajreject}{reject $\times$}{\texttt{show\_account\_passwords} and \texttt{show\_account} are called without reading their documentation, and family members should be looked up in the \textless{}phone\_app\textgreater{}, not the supervisor app.}
\trajpart{Output}
\begin{lstlisting}[style=trajout]
Execution failed. Traceback:
  File "<python-input>", line 11, in <module>
    print(apis.supervisor.show_account())
Exception: No API named 'show_account' found in the supervisor app.
\end{lstlisting}
\end{tcolorbox}
\begin{tcolorbox}[trajstep=trajexpert, unbreakable, title={Step 3 \textperiodcentered{} Expert replacement}]
\trajpart{Expert response}
\begin{lstlisting}[style=trajcode]
print(apis.supervisor.show_account_passwords())
\end{lstlisting}
\trajpart{Output}
\begin{lstlisting}[style=trajout]
[
 {
  "account_name": "<shopping_app>",
  "password": "<password>"
 },
 {
  "account_name": "<file_app>",
  "password": "<password>"
[... 22 more lines ...]
  "account_name": "<todo_app>",
  "password": "<password>"
 },
 {
  "account_name": "<payment_app>",
  "password": "<password>"
 }
]
\end{lstlisting}
\end{tcolorbox}
\end{tcbraster}
\begin{tcbraster}[raster columns=2, raster equal height=rows, raster column skip=6pt,
  raster before skip=5pt, raster after skip=5pt]
\begin{tcolorbox}[trajstep=trajreject, unbreakable, title={Step 4 \textperiodcentered{} Target Model (rejected): States for error-recovery}]
\trajpart{Target Model response}
\begin{lstlisting}[style=trajcode]
# Login to <payment_app>
result = apis.<payment_app>.login(email="<email_address>", password="<password>")
print(f"Login result: {result}")

# Get list of siblings from <phone_app> app
sibling_info = apis.<phone_app>.search_friends(name="<last_name>")
print(f"Siblings info: {sibling_info}")

# Get social feed to see transactions
feed = apis.<payment_app>.show_social_feed()
print(f"Social feed: {feed}")
\end{lstlisting}
\trajjudge{trajreject}{reject $\times$}{\texttt{login}, \texttt{search\_friends} and \texttt{show\_social\_feed} are called without reading their documentation, so their parameters (e.g. \texttt{email} for \texttt{login}) are guesses.}
\trajpart{Output}
\begin{lstlisting}[style=trajout]
Execution failed. Traceback:
  File "<python-input>", line 2, in <module>
    result = apis.<payment_app>.login(email="<email_address>", password="<password>")
Exception: Unexpected parameter 'email' passed to the login API of the <payment_app> app. Allowed parameters are: ['username', 'password']
\end{lstlisting}
\end{tcolorbox}
\begin{tcolorbox}[trajstep=trajexpert, unbreakable, title={Step 4 \textperiodcentered{} Expert replacement}]
\trajpart{Expert response}
\begin{lstlisting}[style=trajcode]
print(apis.api_docs.show_api_descriptions(app_name='<phone_app>'))
\end{lstlisting}
\trajpart{Output}
\begin{lstlisting}[style=trajout]
[
 {
  "name": "show_account",
  "description": "Show your account information. Unlike show_profile, this includes private information."
 },
 {
  "name": "signup",
  "description": "Sign up to create account."
[... 106 more lines ...]
  "name": "send_voice_message",
  "description": "Send a voice message on the given phone number."
 },
 {
  "name": "get_current_date_and_time",
  "description": "Show current date and time."
 }
]
\end{lstlisting}
\end{tcolorbox}
\end{tcbraster}
\begin{tcbraster}[raster columns=2, raster equal height=rows, raster column skip=6pt,
  raster before skip=5pt, raster after skip=5pt]
\begin{tcolorbox}[trajstep=trajreject, unbreakable, title={Step 5 \textperiodcentered{} Target Model (rejected): States for error-recovery}]
\trajpart{Target Model response}
\begin{lstlisting}[style=trajcode]
from datetime import datetime, timedelta

# Get yesterday's date
today = datetime.now()
yesterday = today - timedelta(days=1)
yesterday_str = yesterday.strftime('%Y-%m-%d')

# Login to <payment_app> with my credentials
print("Logging into <payment_app>...")
login_result = apis.<payment_app>.login(email=<email_address>, password='<password>')
print(login_result)

# Search for siblings in contacts to get their usernames/identifiers
print("Searching for siblings in contacts...")
siblings_info = apis.<phone_app>.search_contacts(name='sibling')
print(siblings_info)

# Now search social feed for transactions from yesterday
print("Getting transactions from yesterday's social feed...")
transactions = []
try:
    # Try to get social feed with date filter
    social_feed = apis.<payment_app>.show_social_feed(start_date=yesterday_str, end_date=yesterday_str)
    transactions = social_feed.get('transactions', [])
    print(f"Found {len(transactions)} transactions on the feed from yesterday")
except Exception as e:
    print(f"Error getting social feed: {e}")

# Filter transactions that involve siblings and like them
liked_count = 0
print(f"Found {len(transactions)} transactions from yesterday")
for tx in transactions:
    # Check if any sibling is involved in this transaction
    is_sibling_involved = False
    # Check sender or recipient in transaction
    if transaction.get('sender') and transaction['sender'] in [s['username'] or s['name'] for s in siblings_info]:
        is_sibling_involved = True
    elif transaction.get('receiver') and transaction['receiver'] in [s['username'] or s['name'] for s in siblings_info]:
        is_sibling_involved = True
    
    if is_sibling_involved:
        print(f"Transaction found involving sibling: {tx}")
        apis.<payment_app>.like_transaction(transaction_id=tx)
        liked_count += 1

print(f"Successfully liked {liked_count} transactions from yesterday involving siblings")
\end{lstlisting}
\trajjudge{trajreject}{reject $\times$}{Four APIs are called without their documentation; the email is not quoted (a syntax error) and the loop uses an undefined variable.}
\trajpart{Output}
\begin{lstlisting}[style=trajout]
Execution failed. Traceback:
  File "<python-input>", line 10, in <module>
    login_result = apis.<payment_app>.login(email=<email_address>, password='<password>')
NameError: name '<email_username>' is not defined
\end{lstlisting}
\end{tcolorbox}
\begin{tcolorbox}[trajstep=trajexpert, unbreakable, title={Step 5 \textperiodcentered{} Expert replacement}]
\trajpart{Expert response}
\begin{lstlisting}[style=trajcode]
print(apis.api_docs.show_api_doc(app_name='<phone_app>', api_name='search_contacts'))
\end{lstlisting}
\trajpart{Output}
\begin{lstlisting}[style=trajout]
{
 "app_name": "<phone_app>",
 "api_name": "search_contacts",
 "path": "/<phone_app>/contacts",
 "method": "GET",
 "description": "Search your contact book for relatives' information.",
 "parameters": [
  {
[... 58 more lines ...]
    "created_at": "2019-01-01T00:00:00"
   }
  ],
  "failure": {
   "message": "string"
  }
 }
}
\end{lstlisting}
\end{tcolorbox}
\end{tcbraster}
\begin{tcolorbox}[trajstep=trajaccept, title={Step 6 \textperiodcentered{} Target Model (accepted)}]
\trajpart{Target Model response}
\begin{lstlisting}[style=trajcode]
import datetime
print(apis.api_docs.show_api_doc(app_name='<payment_app>', api_name='login'))
\end{lstlisting}
\trajjudge{trajaccept}{accept $\checkmark$}{Reads the \textless{}payment\_app\textgreater{} \texttt{login} documentation before logging in.}
\trajpart{Output}
\begin{lstlisting}[style=trajout]
{
 "app_name": "<payment_app>",
 "api_name": "login",
 "path": "/<payment_app>/auth/token",
 "method": "POST",
 "description": "Login to your account.",
 "parameters": [
  {
[... 18 more lines ...]
   "access_token": "string",
   "token_type": "string"
  },
  "failure": {
   "message": "string"
  }
 }
}
\end{lstlisting}
\end{tcolorbox}
\begin{tcolorbox}[trajstep=trajaccept, title={Step 7 \textperiodcentered{} Target Model (accepted)}]
\trajpart{Target Model response}
\begin{lstlisting}[style=trajcode]
import datetime
<payment_app>_credentials = {
    'access_token': None
}
response = apis.<payment_app>.login(
    username='<email_address>',
    password='<password>'
)
<payment_app>_credentials['access_token'] = response['access_token']
print("Logged in to <payment_app> successfully!")
print(response)
\end{lstlisting}
\trajjudge{trajaccept}{accept $\checkmark$}{Logs in to the \textless{}payment\_app\textgreater{} with the documented parameters, the user's email and the password from step 3.}
\trajpart{Output}
\begin{lstlisting}[style=trajout]
Logged in to <payment_app> successfully!
{
 "access_token": "<access_token>",
 "token_type": "Bearer"
}
\end{lstlisting}
\end{tcolorbox}
\begin{tcolorbox}[trajstep=trajaccept, title={Step 8 \textperiodcentered{} Target Model (accepted)}]
\trajpart{Target Model response}
\begin{lstlisting}[style=trajcode]
print(apis.api_docs.show_api_doc(app_name='<payment_app>', api_name='show_social_feed'))
\end{lstlisting}
\trajjudge{trajaccept}{accept $\checkmark$}{Reads the \texttt{show\_social\_feed} documentation before calling it.}
\trajpart{Output}
\begin{lstlisting}[style=trajout]
{
 "app_name": "<payment_app>",
 "api_name": "show_social_feed",
 "path": "/<payment_app>/social_feed",
 "method": "GET",
 "description": "Show your social feed (transactions of your friends).",
 "parameters": [
  {
[... 46 more lines ...]
    }
   }
  ],
  "failure": {
   "message": "string"
  }
 }
}
\end{lstlisting}
\end{tcolorbox}
\begin{tcbraster}[raster columns=2, raster equal height=rows, raster column skip=6pt,
  raster before skip=5pt, raster after skip=5pt]
\begin{tcolorbox}[trajstep=trajreject, unbreakable, title={Step 9 \textperiodcentered{} Target Model (rejected): States for error-recovery}]
\trajpart{Target Model response}
\begin{lstlisting}[style=trajcode]
import datetime
# Get current date to determine yesterday
today = datetime.datetime.now()
yesterday = today - datetime.timedelta(days=1)
yesterday_date_str = yesterday.strftime("%Y-%m-%dT%H:%M:%S")
print(f"Today: {today.strftime('%Y-%m-%d')}")
print(f"Yesterday: {yesterday.strftime('%Y-%m-%d')}")
print(f"Yesterday datetime string for filtering: {yesterday_date_str}")

# Now get siblings from <phone_app> app
siblings = []
for page_index in range(0, 5):  # Check multiple pages
    if page_index > 0:
        break  # Will handle page iteration
    page_limit = 20
    
    contacts = apis.<phone_app>.search_contacts(
        access_token='<access_token>',
        query='',
        relationship='sibling',
        page_index=page_index,
        page_limit=page_limit
    )
    print(f"Page {page_index}: {contacts}")
\end{lstlisting}
\trajjudge{trajreject}{reject $\times$}{Calls the \textless{}phone\_app\textgreater{} \texttt{search\_contacts} with the \textless{}payment\_app\textgreater{} access token, without first logging in to the \textless{}phone\_app\textgreater{}.}
\trajpart{Output}
\begin{lstlisting}[style=trajout]
Execution failed. Traceback:
  File "<python-input>", line 17, in <module>
    contacts = apis.<phone_app>.search_contacts(
Exception: Response status code is 401:
{"message":"You are either not authorized to access this <phone_app> API endpoint or your access token is missing, invalid or expired."}
\end{lstlisting}
\end{tcolorbox}
\begin{tcolorbox}[trajstep=trajexpert, unbreakable, title={Step 9 \textperiodcentered{} Expert replacement}]
\trajpart{Expert response}
\begin{lstlisting}[style=trajcode]
print(apis.api_docs.show_api_doc(app_name='<phone_app>', api_name='login'))
\end{lstlisting}
\trajpart{Output}
\begin{lstlisting}[style=trajout]
{
 "app_name": "<phone_app>",
 "api_name": "login",
 "path": "/<phone_app>/auth/token",
 "method": "POST",
 "description": "Login to your account.",
 "parameters": [
  {
[... 18 more lines ...]
   "access_token": "string",
   "token_type": "string"
  },
  "failure": {
   "message": "string"
  }
 }
}
\end{lstlisting}
\end{tcolorbox}
\end{tcbraster}
\begin{tcolorbox}[trajstep=trajaccept, title={Step 10 \textperiodcentered{} Target Model (accepted)}]
\trajpart{Target Model response}
\begin{lstlisting}[style=trajcode]
<phone_app>_credentials = {
    'access_token': None
}
response = apis.<phone_app>.login(
    username='<phone_number>',
    password='<password>'
)
<phone_app>_credentials['access_token'] = response['access_token']
print("Logged in to <phone_app> app successfully!")
print(response)
\end{lstlisting}
\trajjudge{trajaccept}{accept $\checkmark$}{Logs in to the \textless{}phone\_app\textgreater{} with the phone number and password, after reading its \texttt{login} documentation in step 9.}
\trajpart{Output}
\begin{lstlisting}[style=trajout]
Logged in to <phone_app> app successfully!
{
 "access_token": "<access_token>",
 "token_type": "Bearer"
}
\end{lstlisting}
\end{tcolorbox}
\begin{tcolorbox}[trajstep=trajaccept, title={Step 11 \textperiodcentered{} Target Model (accepted)}]
\trajpart{Target Model response}
\begin{lstlisting}[style=trajcode]
siblings = []
page_index = 0
while True:
    response = apis.<phone_app>.search_contacts(
        access_token=<phone_app>_credentials['access_token'],
        relationship='sibling',
        page_index=page_index,
        page_limit=20
    )
    if isinstance(response, list):
        siblings.extend(response)
        if len(response) < 20:
            break
        page_index += 1
    else:
        break

print(f"Found {len(siblings)} siblings")
for sibling in siblings:
    print(f"  - {sibling['first_name']} {sibling['last_name']}, email: {sibling['email']}")
\end{lstlisting}
\trajjudge{trajaccept}{accept $\checkmark$}{Searches the contacts with relationship \texttt{sibling} and handles pagination, using the \textless{}phone\_app\textgreater{} token.}
\trajpart{Output}
\begin{lstlisting}[style=trajout]
Found 3 siblings
  - <sibling_0>, email: <sibling_0_email>
  - <sibling_1>, email: <sibling_1_email>
  - <sibling_2>, email: <sibling_2_email>
\end{lstlisting}
\end{tcolorbox}
\begin{tcolorbox}[trajstep=trajaccept, title={Step 12 \textperiodcentered{} Target Model (accepted)}]
\trajpart{Target Model response}
\begin{lstlisting}[style=trajcode]
import datetime
yesterday_date = (datetime.datetime.now() - datetime.timedelta(days=1)).strftime("%Y-%m-%d")
print(f"Today's date: {datetime.datetime.now().date()}")
print(f"Yesterday's date: {yesterday_date}")
\end{lstlisting}
\trajjudge{trajaccept}{accept $\checkmark$}{Computes yesterday's date programmatically, as the instructions require.}
\trajpart{Output}
\begin{lstlisting}[style=trajout]
Today's date: 2023-05-18
Yesterday's date: 2023-05-17
\end{lstlisting}
\end{tcolorbox}
\begin{tcolorbox}[trajstep=trajaccept, title={Step 13 \textperiodcentered{} Target Model (accepted)}]
\trajpart{Target Model response}
\begin{lstlisting}[style=trajcode]
# Get the social feed to find transactions from yesterday involving siblings
siblings_emails = [
    '<sibling_0_email>',
    '<sibling_1_email>',
    '<sibling_2_email>'
]

# We need to check each page of the social feed to find transactions from yesterday
# Let's start with page 0
page_index = 0
transactions_from_yesterday = []
siblings_transactions = []

while True:
    response = apis.<payment_app>.show_social_feed(
        access_token=<payment_app>_credentials['access_token'],
        page_index=page_index,
        page_limit=20
    )
    
    if not isinstance(response, list) or len(response) == 0:
        break
    
    for txn in response:
        # Check if transaction is from yesterday (created_at should be 2023-05-17)
        if '2023-05-17' in txn.get('created_at', ''):
            # Check if sibling is involved (either sender or receiver)
            sender = txn.get('sender', {})
            receiver = txn.get('receiver', {})
            
            is_sibling_involved = False
            if sender.get('email') in siblings_emails:
                is_sibling_involved = True
            elif receiver.get('email') in siblings_emails:
                is_sibling_involved = True
            
            if is_sibling_involved:
                siblings_transactions.append(txn)
    
    if len(response) < 20:
        break
    page_index += 1

print(f"Found {len(siblings_transactions)} transactions from yesterday involving siblings")
for txn in siblings_transactions:
    print(f"Transaction ID: {txn['transaction_id']}, Sender: {txn['sender']['name']}, Receiver: {txn['receiver']['name']}, Amount: {txn['amount']}")
\end{lstlisting}
\trajjudge{trajaccept}{accept $\checkmark$}{Pages through the social feed and keeps yesterday's transactions whose sender or receiver is a sibling.}
\trajpart{Output}
\begin{lstlisting}[style=trajout]
Found 8 transactions from yesterday involving siblings
Transaction ID: 8223, Sender: <person_0>, Receiver: <sibling_1>, Amount: 15.0
Transaction ID: 8218, Sender: <person_1>, Receiver: <sibling_0>, Amount: 11.0
Transaction ID: 5215, Sender: <sibling_0>, Receiver: <person_2>, Amount: 15.0
Transaction ID: 5224, Sender: <sibling_0>, Receiver: <person_3>, Amount: 146.0
Transaction ID: 8220, Sender: <person_4>, Receiver: <sibling_0>, Amount: 82.0
Transaction ID: 3299, Sender: <person_5>, Receiver: <sibling_0>, Amount: 78.0
Transaction ID: 4243, Sender: <person_6>, Receiver: <sibling_1>, Amount: 14.0
Transaction ID: 6909, Sender: <person_7>, Receiver: <sibling_0>, Amount: 63.0
\end{lstlisting}
\end{tcolorbox}
\begin{tcolorbox}[trajstep=trajaccept, title={Step 14 \textperiodcentered{} Target Model (accepted)}]
\trajpart{Target Model response}
\begin{lstlisting}[style=trajcode]
# Check the like_transaction API documentation
print(apis.api_docs.show_api_doc(app_name='<payment_app>', api_name='like_transaction'))
\end{lstlisting}
\trajjudge{trajaccept}{accept $\checkmark$}{Reads the \texttt{like\_transaction} documentation before calling it.}
\trajpart{Output}
\begin{lstlisting}[style=trajout]
{
 "app_name": "<payment_app>",
 "api_name": "like_transaction",
 "path": "/<payment_app>/transactions/{transaction_id}/likes",
 "method": "POST",
 "description": "Like a transaction.",
 "parameters": [
  {
[... 17 more lines ...]
  "success": {
   "message": "string"
  },
  "failure": {
   "message": "string"
  }
 }
}
\end{lstlisting}
\end{tcolorbox}
\begin{tcolorbox}[trajstep=trajaccept, title={Step 15 \textperiodcentered{} Target Model (accepted)}]
\trajpart{Target Model response}
\begin{lstlisting}[style=trajcode]
# Like all 8 transactions from yesterday involving siblings
target_transaction_ids = [8223, 8218, 5215, 5224, 8220, 3299, 4243, 6909]

for txn_id in target_transaction_ids:
    response = apis.<payment_app>.like_transaction(
        transaction_id=txn_id,
        access_token=<payment_app>_credentials['access_token']
    )
    print(f"Transaction {txn_id}: {response}")
\end{lstlisting}
\trajjudge{trajaccept}{accept $\checkmark$}{Likes all 8 transactions found in step 13 with the documented parameters.}
\trajpart{Output}
\begin{lstlisting}[style=trajout]
Transaction 8223: {'message': 'Liked transaction.'}
Transaction 8218: {'message': 'Liked transaction.'}
Transaction 5215: {'message': 'Liked transaction.'}
Transaction 5224: {'message': 'Liked transaction.'}
Transaction 8220: {'message': 'Liked transaction.'}
Transaction 3299: {'message': 'Liked transaction.'}
Transaction 4243: {'message': 'Liked transaction.'}
Transaction 6909: {'message': 'Liked transaction.'}
\end{lstlisting}
\end{tcolorbox}
\begin{tcolorbox}[trajstep=trajaccept, title={Step 16 \textperiodcentered{} Target Model (accepted)}]
\trajpart{Target Model response}
\begin{lstlisting}[style=trajcode]
apis.supervisor.complete_task()
\end{lstlisting}
\trajjudge{trajaccept}{accept $\checkmark$}{All matching transactions are liked, so completing the task without an answer is correct.}
\trajpart{Output}
\begin{lstlisting}[style=trajout]
Execution successful.
\end{lstlisting}
\end{tcolorbox}

\captionof{figure}{Example \ours trajectory on AppWorld with \texttt{Qwen3.5-9B} target model and \texttt{GLM-5.1} as the expert and judge. The rollout follows target model's generation when the proposed action is accepted by the judge (\textcolor{trajaccept}{green colored steps}), while using expert's generation (\textcolor{trajexpert}{blue colored steps}) when the proposed action is rejected. The rejected steps (\textcolor{trajreject}{red colored}) are used to collect error-recovery supervision.}
\label{fig:appworld_example_1}
\end{center}

\applefootnote{ \textcolor{textgray}{\sffamily Apple and the Apple logo are trademarks of Apple Inc., registered in the U.S. and other countries and regions.}}

\end{document}